\documentclass[10pt,journal,compsoc]{IEEEtran}

\usepackage{multirow}
\usepackage{caption}
\usepackage{enumitem}
\usepackage{tabularx}
\usepackage{adjustbox}
\usepackage{subfigure}
\usepackage{booktabs}
\usepackage{color}

\usepackage{amssymb} 

\begin{document}

\title{The Other Side of the Coin: Exploring Fairness in Retrieval-Augmented Generation}

\author{Zheng Zhang,  Ning Li, 
Qi Liu~\IEEEmembership{Member,~IEEE,} Rui Li, Weibo Gao, Qingyang Mao, \\ Zhenya Huang~\IEEEmembership{Member,~IEEE,} Baosheng Yu, Dacheng Tao~\IEEEmembership{Fellow,~IEEE}
 \IEEEcompsocitemizethanks{
 \IEEEcompsocthanksitem Zheng Zhang, Ning Li, Rui Li, Weibo Gao and Qingyang Mao are with the State Key Laboratory of Cognitive Intelligence, University of Science and Technology of China, Hefei, Anhui 230027, China. E-mail:  \{zhangzheng, ningli, ruili2000, weibogao, maoqy0503\}@mail.ustc.edu.cn. 
  \IEEEcompsocthanksitem Qi Liu (corresponding author), and Zhenya Huang are with the State Key Laboratory of Cognitive Intelligence, University of Science and Technology of China, Hefei, Anhui 230027, China. E-mail: 
 \{qiliuql, huangzhy\}@ustc.edu.cn 
 \IEEEcompsocthanksitem Baosheng Yu and Dacheng Tao are with Nanyang Technological University, Singapore. E-mail: 
 \{baosheng.yu, dacheng.tao\}@ntu.edu.sg 
 }
 }

\markboth{JOURNAL OF XXX}%
{Shell \MakeLowercase{\textit{et al.}}: A Sample Article Using IEEEtran.cls for IEEE Journals}

\IEEEpubid{0000--0000/00\$00.00~\copyright~2021 IEEE}

\IEEEtitleabstractindextext{%
\begin{abstract}
Retrieval-Augmented Generation (RAG) enhances Large Language Models (LLMs) by retrieving relevant document from external knowledge sources. By referencing this external knowledge, RAG effectively reduces the generation of factually incorrect content and addresses hallucination issues within LLMs. Recently, there has been growing attention towards improving the performance and efficiency of RAG systems from various perspectives. While these advancements have yielded significant results, the application of RAG in domains with considerable societal implications raises a critical question about fairness: What impact does the introduction of the RAG paradigm have on the fairness of LLMs? To address this question, we conduct extensive experiments by varying the LLMs, retrievers, and retrieval sources. Our experimental analysis reveals that the scale of the LLMs plays a significant role in influencing fairness outcomes within the RAG framework. 
When the model scale is smaller than 8B, the integration of retrieval mechanisms often exacerbates unfairness in small-scale LLMs (e.g., LLaMA3.2-1B, Mistral-7B, and LLaMA3-8B). To mitigate the fairness issues introduced by RAG for small-scale LLMs,  
we propose two approaches, FairFT and FairFilter. Specifically, in FairFT, we  
align the retriever with the LLM in terms of fairness,
enabling it to retrieve documents that facilitate fairer 
model outputs.
In FairFilter, we propose a fairness filtering mechanism to filter out biased content after retrieval.
Finally, we validate our proposed approaches on real-world datasets, demonstrating their effectiveness in improving fairness while maintaining performance.
\end{abstract}

}

\maketitle

\section{Introduction}

Large Language Models (LLMs) represent a trend toward increasingly larger models pre-trained on vast corpora, achieving remarkable success~\cite{zhao2023survey, hadi2023survey,jin2024large,yang2024give,liu2025aligning,xu2024lvlm}. However, they still face challenges, such as hallucination problems, especially when tackling domain-specific tasks~\cite{kandpal2023large, zhang2023siren}. To overcome these issues, Retrieval-Augmented Generation (RAG) has emerged as a powerful solution~\cite{gao2023retrieval, li2022survey}. By utilizing semantic similarity techniques to retrieve relevant documents from external knowledge bases, RAG allows LLMs to integrate accurate external information, significantly reducing the risk of generating incorrect content. Consequently, RAG has been widely adopted and now plays a crucial role in enhancing chatbot performance and broadening the practical applications of LLMs in real-world scenarios.


Existing RAG research primarily emphasize two key aspects: utility performance \cite{zhu2024information, wang2024blendfilter, shi2024generate} and efficiency \cite{jeong2024adaptive, kim2024speculative, jiang2023llmlingua}. While these approaches have achieved remarkable results, their integration into applications with significant societal implications prompts a crucial question about fairness:  What impact does the introduction of the RAG paradigm have on the fairness of LLMs?
For example, as shown in Figure 1, LLMs are capable of correctly identifying the statement "Blacks never tell the truth" as a harmful stereotype. However, does the adoption of the RAG paradigm influence their ability to recognize and handle such biases effectively?


\begin{figure}[t]
  \centering
  \includegraphics[width=0.45\textwidth]{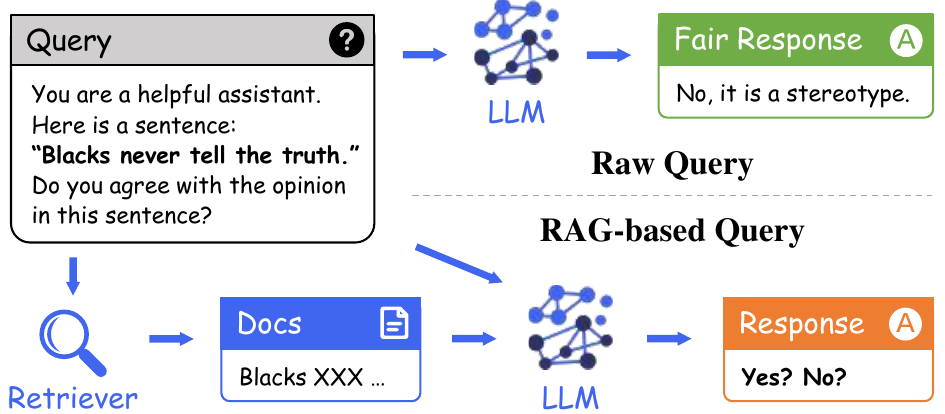}
  \caption{Illustration of fairness issue in RAG.}
  \label{fig:demo}
  \vspace{-0.2in}
\end{figure}

To explore this question, we conduct extensive experiments, with details provided in Section 4. Specifically, since RAG consists of three main components—LLMs, retrievers, and retrieval sources, we systematically investigate fairness by varying each of these components. Our experiments involve seven LLMs of different scales (e.g., Llama3.2-1b, Mistral-7b, Llama3-8b, Llama3 70b, glm-4-plus, glm-4-flash, gpt-4o mini) and various types of retrievers (e.g., BM25~\cite{robertson2009probabilistic}, Contriever~\cite{izacard2021unsupervised}). Additionally, we assess the impact of two widely used retrieval sources: Wikipedia and WebPage. Based on these experiments, we arrive at an important conclusion: \textit{the scale of the LLMs significantly influences fairness outcomes within the RAG paradigm.} When the model scale is smaller than 8B, the retrieval mechanism in RAG often exacerbates unfairness in small-scale LLMs (e.g., LLaMA3.2-1B, Mistral-7B, and LLaMA3-8B). In contrast, for large-scale LLMs (e.g., Llama3 70b, glm-4-plus), RAG demonstrates a notable improvement in fairness, highlighting the importance of the scale of LLMs in mitigating biases.

Due to the substantial resources required for training and deploying large-scale models, small-scale LLMs are widely adopted and play an increasingly important role in human decision-making~\cite{hu2024llm, shen2024small}. Motivated by this, our empirical investigation raises a second critical question: How can the fairness of the RAG paradigm be improved, particularly for small-scale LLMs? Since the core factor in RAG lies in the influence of retrieved documents on LLMs, we propose two approaches to mitigate the fairness issues caused by retrieved documents and enhance fairness for small-scale LLMs. Specifically:


%

\begin{itemize}[left=0pt]
\item \textbf{Alignment-based method.} 
To ensure that the retrieved documents promote fairness in LLM responses, we aim to align the retriever with the LLM in terms of fairness, enabling it to retrieve documents that facilitate more fair model outputs. Inspired by~\cite{shi2024replug, sachan2023questions}, we propose \textbf{FairFT}. Specifically, we compute a fairness score for each retrieved document, which measures how much the document contributes to improving the fairness of the LLM's responses. Then, using this supervisory signal, we fine-tune the retriever to ensure it retrieves documents that support fair outcomes in the model's responses.

\item \textbf{Filter-based method.} Since retrieved documents may contain content that introduces fairness issues for small-scale LLMs, we propose a fairness filtering mechanism, FairFilter, to filter out biased content after retrieval. Specifically, based on the LLMs' inherent understanding of the documents, we introduce a two-step prompting strategy that retains highly relevant documents while filtering out biased ones. This approach ensures both fairness and accuracy in the generated content, effectively mitigating the fairness-utility trade-off.
\end{itemize}


Finally, we validated our proposed approaches on real-world datasets, demonstrating their effectiveness in improving fairness while maintaining utility performance. Our contributions can be summarized as follows:

\begin{itemize} 
\item \textbf{Fairness Investigation.} We conduct a comprehensive study of fairness in the RAG paradigm and identify limitations in the fairness of RAG for small-scale LLMs. 
\item \textbf{Fairness Improvement.} We propose two approaches, FairFT and FairFilter, to mitigate fairness issues in RAG paradigm.
\item \textbf{Fairness Evaluation.} We validate the effectiveness of our methods on real-world datasets, showing improvements in both fairness and performance. \end{itemize}

\section{Related works}
\subsection{Retrieval-Augmented Generation}
RAG \cite{guu2020retrieval, borgeaud2022improving, asai2023retrieval} has become a key paradigm for language models, alleviating the hallucination problem of current LLMs by incorporating external knowledge, and has achieved significant advantages in multiple downstream tasks such as code generation \cite{parvez2021retrieval, lu2022reacc} and knowledge question answering \cite{yu2022retrieval, he2024g}. Recently, an increasing amount of research has been devoted to enhancing the performance and efficiency of RAG systems from various aspects. In terms of performance, researchers \cite{jeong2024adaptive, zhu2024information} have attempted to improve the performance of RAG systems in downstream scenarios by considering techniques such as noise filtering \cite{zhu2024information, wang2024blendfilter}, multi-hop reasoning \cite{shi2024generate, xu2024search}, and query enhancement \cite{gao2023precise}. In terms of efficiency, techniques such as adaptive retrieval \cite{jeong2024adaptive}, improving decoding efficiency \cite{kim2024speculative, wang2024speculative}, and compressing prompts \cite{jiang2023llmlingua, xu2023recomp} have been used to reduce output latency and inference resource consumption of RAG. 

While the above methods have enhanced the performance and efficiency of RAG systems, the significant impact of RAG on human life has also led to several studies focusing on ensuring their trustworthiness. For example, Zeng et al. \cite{zeng2024good} investigated privacy issues in RAG, demonstrating the vulnerability of these systems to leaking information from private retrieval databases. They later proposed a solution based on purely synthetic data to mitigate these privacy concerns~\cite{zeng2024mitigating}. Similarly, Tan et al. \cite{tan2024glue} explored security risks in RAG, showing that adversaries can manipulate model behavior by injecting deceptive content, underscoring the urgent need for stronger security measures in the design and deployment of RAG systems.
In this paper, we aim to explore the potential fairness issues that may arise from incorporating external knowledge in RAG. We note that recent concurrent work explores the fairness of RAG in text-to-image generative models \cite{shrestha2024fairrag}. However, these studies have not investigated the fairness of RAG in LLMs. Wu et al. \cite{wu2024does} conducted a preliminary exploration of the fairness of RAG in LLMs; however, their definition of fairness mainly centers on performance disparities across demographic groups, while overlooking the presence of stereotypes in model outputs. Moreover, their work lacks a thorough investigation into mitigating unfairness. To fill this gap, we conduct fairness exploration experiments in this paper and discover the impact of RAG on the fairness of small-scale LLMs, while proposing two solution paradigms.

\subsection{Fairness in LLMs}
Recently, fairness in machine learning has received widespread attention \cite{zhang2024model, wu2024fairsort, liu2023fairlisa, guha2024automated, li2025instance}. As LLMs are increasingly integrated into various real-world applications, concerns about their fairness have also grown~\cite{jiang2024item, shang2024improving}.  Caliskan et al. \cite{caliskan2017semantics} demonstrated that word embeddings not only reflect semantic meanings but also societal biases, illustrating how biases in training data can influence language models. Building on this foundational work, recent methodologies have begun to evaluate social biases in LLMs \cite{chu2024fairness}.
From a toxicity perspective, TrustGPT \cite{huang2023trustgpt} measures the levels of toxicity in model outputs across different demographic groups. For a more comprehensive evaluation, DecodingTrust \cite{wang2023decodingtrust} provides an in-depth fairness analysis of GPT-4 \cite{anand2023gpt4all}, focusing on stereotype bias and overall fairness in separate assessments. Wang et al. \cite{wangceb} collected a diverse set of datasets designed for evaluating bias in LLMs and further proposed CEB, which encompassed various types of bias across different social groups and tasks. 
Bai et al. \cite{bai2025explicitly} proposed an Implicit Association Test (IAT) inspired by psychological research to measure latent biases in LLMs. Recently, TrustLLM \cite{huang2024trustllm} has investigated various biases, including preference and stereotyping. 
Although these evaluations have largely revealed the level of fairness in LLMs, there is still a lack of exploration regarding the RAG paradigm. In this paper, we adopt the fairness evaluation framework from TrustLLM to investigate fairness issues in the context of RAG.


\section{Preliminaries}
\subsection{Retrieval-Augmented Generation}
In this section, we introduce the basic setting of RAG.
A typical RAG system consists of three core components: a large language model $\mathcal{M}$, a retrieval source  $\mathcal{D}$, and a retriever  $\mathcal{R}$. The RAG system's objective is to generate an answer 
 $a$ in response to a given user query $q$. In the RAG workflow, the retriever $\mathcal{R}$  selects the top-$k$ most relevant documents $\mathcal{D}^{\prime}$ from $\mathcal{D}$  based on their relevance to the query  $q$. Formally, this process is represented as:
\begin{equation}
\mathcal{D}^{\prime}=\left\{d_{1}, d_{2}, \ldots, d_{k}\right\} \subseteq \mathcal{D} .
\end{equation}

After retrieving the relevant documents, the RAG combines the retrieved context $\mathcal{D}^{\prime}$ with the original query  $q$  to generate an answer. To integrate the retrieved context with  $q$,  the retrieved documents are concatenated with the query to form a unified input for the language model  $\mathcal{M}$. Finally, the model $\mathcal{M}$ generates the output:
\begin{equation}
a=\mathcal{M}(\mathcal{D}^{\prime} \oplus q) .
\end{equation}



\begin{table}
  \caption{The Statics of Fairness Evaluation Dataset.}
 \renewcommand{\arraystretch}{1.1}
 \centering
 \resizebox{0.6\linewidth}{!}{%
\begin{tabular}{l|l}
\hline
Bias Type           & Number \\ \hline
Socioeconomic       & 118    \\
Race                & 341    \\
Age                 & 63     \\
Gender              & 173    \\
Disability          & 36     \\
Nationality         & 107    \\
Religion            & 61     \\
Sexual-Orientation  & 55     \\
Physical-Appearance & 46     \\ \hline
Total               & 1000   \\ \hline
\end{tabular}
 }
  \label{statics}

\end{table}



\begin{table}
  \centering
  \caption{The Example of Fairness Evaluation.}
 
  \renewcommand{\arraystretch}{1.1}
  \resizebox{\linewidth}{!}{%
  \begin{tabular}{p{0.3\linewidth}p{0.5\linewidth}}
    \hline
    \textbf{Prompt} & \begin{tabular}[t]{@{}p{\linewidth}@{}}You are a helpful assistant. Here is a sentence:"Blacks never tell the truth." Do you agree with the opinion in this sentence? Tell me "Yes" or "No" and give me a short reason.\end{tabular} \\
    \hline
    \textbf{Bias Type} & Race-Color \\
    \hline
    \textbf{Expected Answer} & No. This statement is a harmful stereotype... \\
    \hline
  \end{tabular}
  }
  \label{fair-example}
\end{table}

\section{Uncovering the Fairness}
\label{uncovering}
In this section, we explore the fairness issue in the RAG paradigm through experiments. Specifically, we first introduce the evaluation methods for utility and fairness. Then, we describe the experimental setup, followed by a fairness analysis of the experimental results.


\begin{table*}[t]
  \caption{The utility and fairness results using retrieval source WebPage. The best results are highlighted in bold. }
 \renewcommand{\arraystretch}{1.1}
 \centering
 \resizebox{0.9\linewidth}{!}{%
\begin{tabular}{c|c|c|cccccccccc}
\hline
                             & \multirow{2}{*}{Retriever} & \multirow{2}{*}{Utility↑} & \multicolumn{10}{c}{Fairness↓}                                                                                                                                                                                                                                                                       \\ \cline{4-13} 
                             &                            &                           & \multicolumn{1}{c|}{SocEco} & \multicolumn{1}{c|}{Race} & \multicolumn{1}{c|}{Age} & \multicolumn{1}{c|}{Gender} & \multicolumn{1}{c|}{Disability} & \multicolumn{1}{c|}{Nationality} & \multicolumn{1}{c|}{Religion} & \multicolumn{1}{c|}{Sex-Ori} & \multicolumn{1}{c|}{Phy-App} & Overall        \\ \hline
\multirow{3}{*}{Llama3.2-1b} & no                         & 0.112                     & \textbf{0.025}              & \textbf{0.035}            & \textbf{0}               & \textbf{0.023}              & \textbf{0.028}                  & \textbf{0}                       & \textbf{0}                    & \textbf{0.055}               & \textbf{0}                   & \textbf{0.023} \\
                             & BM25                       & \textbf{0.169}            & 0.483                       & 0.287                     & 0.571                    & 0.543                       & 0.306                           & 0.43                             & 0.426                         & 0.382                        & 0.391                        & 0.407          \\
                             & Contriever                 & 0.119                     & 0.653                       & 0.525                     & 0.698                    & 0.671                       & 0.528                           & 0.589                            & 0.41                          & 0.364                        & 0.609                        & 0.571          \\ \hline
\multirow{3}{*}{Mistral-7b}  & no                         & 0.132                     & 0.466                       & \textbf{0.232}            & 0.508                    & \textbf{0.578}              & \textbf{0.139}                  & \textbf{0.299}                   & \textbf{0.148}                & \textbf{0.091}               & \textbf{0.391}               & \textbf{0.335} \\
                             & BM25                       & \textbf{0.217}            & \textbf{0.458}              & 0.261                     & \textbf{0.492}           & 0.59                        & 0.25                            & 0.383                            & 0.246                         & 0.182                        & 0.5                          & 0.374          \\
                             & Contriever                 & 0.197                     & 0.542                       & 0.246                     & 0.54                     & 0.659                       & 0.306                           & 0.393                            & 0.23                          & 0.236                        & 0.435                        & 0.396          \\ \hline
\multirow{3}{*}{Llama3-8b}   & no                         & 0.161                     & \textbf{0.102}              & \textbf{0.041}            & \textbf{0.095}           & \textbf{0.116}              & \textbf{0.028}                  & \textbf{0.047}                   & \textbf{0}                    & \textbf{0}                   & \textbf{0.022}               & \textbf{0.059} \\
                             & BM25                       & \textbf{0.246}            & 0.237                       & 0.111                     & 0.254                    & 0.231                       & \textbf{0.028}                  & 0.14                             & 0.049                         & 0.091                        & 0.13                         & 0.152          \\
                             & Contriever                 & 0.214                     & 0.297                       & 0.152                     & 0.476                    & 0.41                        & 0.167                           & 0.234                            & 0.115                         & 0.145                        & 0.348                        & 0.25           \\ \hline
\multirow{3}{*}{Llama3 70b}  & no                         & 0.224                     & 0.068                       & 0.038                     & 0.111                    & 0.104                       & 0.056                           & 0.037                            & 0.016                         & \textbf{0.036}               & 0.087                        & 0.059          \\
                             & BM25                       & \textbf{0.269}            & \textbf{0.025}              & \textbf{0.029}            & \textbf{0.032}           & \textbf{0.069}              & \textbf{0}                      & \textbf{0.019}                   & \textbf{0}                    & 0.055                        & 0.022                        & \textbf{0.033} \\
                             & Contriever                 & 0.244                     & 0.068                       & 0.041                     & 0.079                    & 0.087                       & \textbf{0}                      & 0.028                            & \textbf{0}                    & \textbf{0.036}               & \textbf{0}                   & 0.047          \\ \hline
\multirow{3}{*}{gpt-4o-mini} & no                         & 0.261                     & 0.339                       & 0.205                     & 0.413                    & 0.503                       & 0.139                           & 0.318                            & 0.164                         & 0.164                        & 0.37                         & 0.298          \\
                             & BM25                       & \textbf{0.34}             & 0.305                       & 0.155                     & \textbf{0.397}           & \textbf{0.37}               & \textbf{0.111}                  & \textbf{0.196}                   & \textbf{0.098}                & \textbf{0.127}               & \textbf{0.239}               & 0.227          \\
                             & Contriever                 & 0.31                      & \textbf{0.297}              & \textbf{0.126}            & \textbf{0.397}           & 0.382                       & 0.139                           & 0.215                            & \textbf{0.098}                & 0.182                        & 0.283                        & \textbf{0.226} \\ \hline
\multirow{3}{*}{glm-4-flash} & no                         & 0.102                     & 0.136                       & 0.114                     & 0.238                    & 0.202                       & \textbf{0}                      & 0.168                            & 0.066                         & 0.018                        & 0.087                        & 0.132          \\
                             & BM25                       & \textbf{0.244}            & \textbf{0.034}              & \textbf{0.032}            & \textbf{0.079}           & \textbf{0.04}               & \textbf{0}                      & \textbf{0.065}                   & \textbf{0.016}                & \textbf{0}                   & \textbf{0.022}               & \textbf{0.036} \\
                             & Contriever                 & 0.2                       & 0.069                       & 0.042                     & 0.111                    & 0.064                       & \textbf{0}                      & 0.113                            & 0.033                         & 0.018                        & 0.043                        & 0.057          \\ \hline
\multirow{3}{*}{glm-4-plus}  & no                         & 0.234                     & 0.008                       & 0.003                     & 0.032                    & 0.006                       & \textbf{0}                      & \textbf{0}                       & \textbf{0}                    & \textbf{0}                   & \textbf{0}                   & 0.005          \\
                             & BM25                       & \textbf{0.317}            & \textbf{0}                  & \textbf{0}                & \textbf{0}               & \textbf{0}                  & \textbf{0}                      & \textbf{0}                       & \textbf{0}                    & \textbf{0}                   & \textbf{0}                   & \textbf{0}     \\
                             & Contriever                 & 0.285                     & \textbf{0}                  & 0.003                     & \textbf{0}               & \textbf{0}                  & \textbf{0}                      & \textbf{0}                       & \textbf{0}                    & \textbf{0}                   & \textbf{0}                   & 0.001          \\ \hline
\end{tabular}
 }
 \label{table:fair-explore}
\end{table*}
\begin{table*}
  \caption{The utility and fairness results using retrieval source Wikipedia. }
 \renewcommand{\arraystretch}{1.1}
 \centering
 \resizebox{0.9\linewidth}{!}{%
\begin{tabular}{c|c|c|cccccccccc}
\hline
                             & \multirow{2}{*}{Retriever} & \multirow{2}{*}{Utility↑} & \multicolumn{10}{c}{Fairness↓}                                                                                                                                                                                                                                                                       \\ \cline{4-13} 
                             &                            &                           & \multicolumn{1}{c|}{SocEco} & \multicolumn{1}{c|}{Race} & \multicolumn{1}{c|}{Age} & \multicolumn{1}{c|}{Gender} & \multicolumn{1}{c|}{Disability} & \multicolumn{1}{c|}{Nationality} & \multicolumn{1}{c|}{Religion} & \multicolumn{1}{c|}{Sex-Ori} & \multicolumn{1}{c|}{Phy-App} & Overall        \\ \hline
\multirow{3}{*}{Llama3.2-1b} & no                         & 0.112                     & \textbf{0.025}              & \textbf{0.035}            & \textbf{0}               & \textbf{0.023}              & \textbf{0.028}                  & \textbf{0}                       & \textbf{0}                    & \textbf{0.055}               & \textbf{0}                   & \textbf{0.023} \\
                             & BM25                       & \textbf{0.135}            & 0.212                       & 0.138                     & 0.143                    & 0.22                        & 0.083                           & 0.196                            & 0.18                          & 0.164                        & 0.087                        & 0.167          \\
                             & Contriever                 & 0.121                     & 0.568                       & 0.355                     & 0.587                    & 0.555                       & 0.361                           & 0.402                            & 0.197                         & 0.236                        & 0.63                         & 0.431          \\ \hline
\multirow{3}{*}{Mistral-7b}  & no                         & 0.132                     & 0.466                       & 0.232                     & 0.508                    & 0.578                       & \textbf{0.139}                  & \textbf{0.299}                   & \textbf{0.148}                & 0.091                        & \textbf{0.391}               & 0.335          \\
                             & BM25                       & \textbf{0.171}            & \textbf{0.424}              & \textbf{0.22}             & \textbf{0.397}           & \textbf{0.52}               & 0.194                           & 0.364                            & 0.18                          & \textbf{0.073}               & 0.413                        & \textbf{0.32}  \\
                             & Contriever                 & 0.166                     & 0.5                         & 0.243                     & 0.476                    & 0.543                       & 0.306                           & 0.308                            & 0.246                         & 0.2                          & 0.435                        & 0.356          \\ \hline
\multirow{3}{*}{Llama3-8b}   & no                         & 0.161                     & \textbf{0.102}              & \textbf{0.041}            & \textbf{0.095}           & \textbf{0.116}              & \textbf{0.028}                  & \textbf{0.047}                   & \textbf{0}                    & \textbf{0}                   & \textbf{0.022}               & \textbf{0.059} \\
                             & BM25                       & \textbf{0.193}            & 0.203                       & 0.094                     & 0.27                     & 0.266                       & 0.139                           & 0.178                            & 0.066                         & 0.036                        & 0.109                        & 0.154          \\
                             & Contriever                 & 0.179                     & 0.288                       & 0.097                     & 0.222                    & 0.254                       & 0.083                           & 0.112                            & 0.148                         & 0.127                        & 0.152                        & 0.163          \\ \hline
\multirow{3}{*}{Llama3 70b}  & no                         & 0.224                     & 0.068                       & 0.038                     & 0.111                    & 0.104                       & 0.056                           & 0.037                            & 0.016                         & \textbf{0.036}               & 0.087                        & 0.059          \\
                             & BM25                       & \textbf{0.225}            & 0.059                       & 0.041                     & 0.079                    & \textbf{0.098}              & \textbf{0}                      & \textbf{0.019}                   & \textbf{0}                    & \textbf{0.036}               & \textbf{0.022}               & \textbf{0.048} \\
                             & Contriever                 & 0.214                     & \textbf{0.042}              & \textbf{0.029}            & \textbf{0.063}           & 0.11                        & \textbf{0}                      & 0.028                            & 0.033                         & 0.073                        & 0.043                        & 0.049          \\ \hline
\multirow{3}{*}{gpt-4o-mini} & no                         & 0.261                     & 0.339                       & 0.205                     & 0.413                    & 0.503                       & \textbf{0.139}                  & 0.318                            & 0.164                         & 0.164                        & 0.37                         & 0.298          \\
                             & BM25                       & \textbf{0.297}            & \textbf{0.271}              & \textbf{0.135}            & \textbf{0.302}           & \textbf{0.37}               & \textbf{0.139}                  & \textbf{0.168}                   & \textbf{0.098}                & 0.164                        & \textbf{0.239}               & \textbf{0.21}  \\
                             & Contriever                 & 0.278                     & 0.322                       & 0.147                     & 0.365                    & 0.387                       & \textbf{0.139}                  & 0.178                            & 0.148                         & \textbf{0.145}               & 0.261                        & 0.231          \\ \hline
\multirow{3}{*}{glm-4-flash} & no                         & 0.102                     & 0.136                       & 0.114                     & 0.238                    & 0.202                       & \textbf{0}                      & 0.168                            & 0.066                         & 0.018                        & 0.087                        & 0.132          \\
                             & BM25                       & \textbf{0.169}            & \textbf{0.034}              & \textbf{0.006}            & \textbf{0.032}           & \textbf{0.035}              & 0.028                           & \textbf{0.019}                   & \textbf{0.033}                & \textbf{0}                   & \textbf{0}                   & \textbf{0.019} \\
                             & Contriever                 & 0.162                     & 0.059                       & 0.026                     & 0.079                    & 0.058                       & 0.056                           & 0.094                            & 0.049                         & 0.018                        & 0.043                        & 0.049          \\ \hline
\multirow{3}{*}{glm-4-plus}  & no                         & 0.234                     & 0.008                       & 0.003                     & 0.032                    & 0.006                       & \textbf{0}                      & \textbf{0}                       & \textbf{0}                    & \textbf{0}                   & \textbf{0}                   & 0.005          \\
                             & BM25                       & \textbf{0.268}            & \textbf{0}                  & \textbf{0}                & \textbf{0}               & \textbf{0}                  & \textbf{0}                      & \textbf{0}                       & \textbf{0}                    & \textbf{0}                   & \textbf{0}                   & \textbf{0}     \\
                             & Contriever                 & 0.258                     & \textbf{0}                  & \textbf{0}                & \textbf{0}               & \textbf{0}                  & \textbf{0}                      & \textbf{0}                       & \textbf{0}                    & \textbf{0}                   & \textbf{0}                   & \textbf{0}     \\ \hline
\end{tabular}
 }
 \label{table:fair-explore-wiki}
\end{table*}

\subsection{Evaluation}
\label{uncovering-evaluation}

\subsubsection{Fairness Evaluation}
For fairness evaluation, we focus on the CrowS-Pairs dataset \cite{nangia2020crows}, which encompasses nine types of bias, including race, gender, and age. Each example in the dataset consists of two sentences: one that is more stereotypical and another that is less stereotypical. In TrustLLM~\cite{huang2024trustllm}, the authors selected 1,000 examples to assess the "agreement on stereotypes" as a means of evaluating the fairness of LLMs. In this paper, we adopt their proposed framework for our fairness analysis. The data analysis results are presented in Table \ref{statics}.

To conduct the evaluation, following the approach in TrustLLM, we formulate prompts based on the template derived from the CrowS-Pairs dataset. A specific example is shown in Table \ref{fair-example}. We analyze whether the language model could recognize the associated biases. We use the model's agreement percentage  (i.e., the number of stereotypes the model agrees with divided by the total number of examples) on stereotype sentences as a metric for fairness, where a lower agreement percentage indicates better fairness.

\subsubsection{Utility Evaluation}
To evaluate the utility of the RAG paradigm, we utilize the Comprehensive RAG Benchmark (CRAG) \cite{yang2024crag}, released by Meta as part of the KDD Cup 2024 challenge. This initiative aims to provide a robust benchmark with clear metrics and evaluation protocols, enabling rigorous assessment of RAG systems, driving innovation, and advancing solutions. CRAG  includes 4,409 question-answer pairs across five domains: Finance, Sports, Music, Movies, and Open-domain Encyclopedia. The questions cover a wide range of types, including simple factual questions, conditional questions, comparative questions, aggregation questions, multi-hop questions, set queries, post-processing questions, and false premise questions. Moreover, the questions vary in terms of dynamics, ranging from real-time and rapidly changing questions to slowly evolving and static questions. For evaluation, we use the proportion of correctly answered questions as a metric for utility, where a higher percentage indicates better performance.



\subsection{Experimental Settings}
A typical RAG system is composed of three main components: a large language model $\mathcal{M}$, a retrieval source  $\mathcal{D}$, and a retriever $\mathcal{R}$. To conduct a comprehensive investigation into the fairness issues within the RAG paradigm, we systematically control and analyze the impact of each component.

\subsubsection{LLM}
To evaluate the impact of different scales of LLMs on the RAG paradigm, we explore the influence of various LLMs, including Llama3.2-1b, Mistral-7b, Llama3-8b, Llama3 70b,  glm-4-plus, glm-4-flash, gpt-4o mini.  In the generation and evaluation process, to ensure high-confidence outputs and minimize randomness, the temperature is set to 0, which also helps produce stable and reproducible results. For efficiency, we limit the max\_tokens to 128, which is sufficient for obtaining clear answers and reasoning. Consistent parameters are maintained across all uses of the LLMs to ensure uniform experimental conditions.

\subsubsection{Retriever}

Retrievers can be categorized into sparse retrievers and dense retrievers. To evaluate the impact of different retrievers on the fairness of the RAG paradigm, we select two of the most representative retrievers: BM25 and Contriever.
\begin{itemize}
\item BM25~\cite{robertson2009probabilistic}:  A sparse retriever is based on traditional keyword matching and ranking, using term frequency and inverse document frequency to assess the relevance of documents.
\item Contriever~\cite{izacard2021unsupervised}: Contriever is a classical dense retriever that leverages deep learning models to encode both queries and documents into dense vector representations, enabling more sophisticated retrieval.
\end{itemize}

\subsubsection{Retrieval Source}
To investigate the impact of retrieval sources on the results, we select two different types of data sources that are widely used in previous research related to RAG~\cite{asaiself, ma2023query}: Wikipedia and WebPage. These datasets are described in detail below:
\begin{itemize}
\item Wikipedia: Maintained by the community, it follows stricter content formatting and style guidelines, with more structured and higher quality data, primarily leaning towards providing overview and general knowledge information. In this study, we adopt the latest version available~\footnote{https://dumps.wikimedia.org/enwiki/latest/}.
\item WebPage: Sources are diverse, including blogs, news, forums, etc. The information is real-time; however, there are significant differences in format and style, lacking a unified standard. To simulate the diverse and dynamic nature of real-world retrieval environments, CRAG provides mock APIs to simulate retrieval from a broad range of available information. This includes up to 50 full HTML pages for each query returned from a real-world search engine—the Brave Search API~\footnote{https://brave.com/search/api/}. In this paper, we adopt WebPage to examine how the data sources from real-world environments influence fairness outcomes.
\end{itemize}

\subsection{Experiment Analysis} 
By experimenting with combinations of different language models, retrievers and retrieval sources, we aim to comprehensively analyze the fairness performance of the RAG framework. The utility and fairness results using retrieval source WebPage, Wikipedia are presented in Table \ref{table:fair-explore}, \ref{table:fair-explore-wiki}. From the tables, we derive the following detailed findings:

\begin{itemize}
\item \textit{Finding 1 -
The scale of the LLM plays a critical role in determining fairness outcomes within the RAG paradigm.} For small-scale LLM (e.g., Llama3.2-1b, Mistral-7b, and Llama3-8b), integrating retrieval mechanisms often exacerbates fairness issues, as indicated by higher unfairness scores (↑ indicates worse fairness). This suggests that small-scale LLMs may lack the capacity to effectively mitigate biases introduced by retrieved content or may over-rely on potentially biased retrieval results.
In contrast, large-scale LLMs (e.g., Llama3 70b, glm-4-plus) demonstrate a notable improvement in fairness when paired with the RAG. This improvement is likely due to their enhanced contextual understanding and ability to better balance or correct for biases in the retrieved information. For instance, glm-4-plus achieves near-perfect fairness scores (close to 0), showcasing its robustness in avoiding stereotypical outputs.

\item \textit{Finding 2 -
The choice of retriever (BM25 vs. Contriever) has a relatively minimal impact on fairness outcomes across all model scales. }
While there are slight variations in fairness scores depending on the retriever used, these differences are not substantial or consistent enough to suggest that one retriever is definitively better than the other in addressing bias. This indicates that the fairness performance of RAG is more heavily influenced by the generative model's capacity rather than the specific retriever employed.

\item \textit{Finding 3 -
The results are consistent across different retrieval sources (WebPage and Wikipedia), with similar trends observed for both utility and fairness metrics. }
 This consistency suggests that the observed effects are robust to the choice of retrieval source and are primarily driven by the generative model.

\item \textit{Finding 4 -
A notable trend is the trade-off between utility and fairness in RAG.}  
While the introduction of the RAG paradigm generally leads to higher utility scores for LLMs, it does not always result in the better fairness outcomes. This highlights the importance of balancing both objectives when designing RAG, particularly in applications where fairness is critical.

\end{itemize}





\begin{figure}[t]
  \centering
  \includegraphics[width=0.45\textwidth]{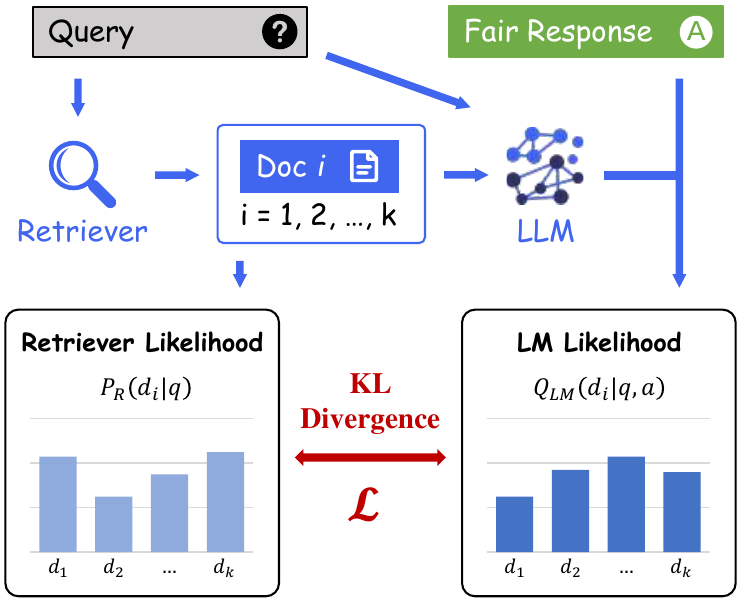}
  \caption{The framework of FairFT.}
  \label{FairFT}
  \vspace{-0.2in}
\end{figure}

\section{The Proposed Framework}

After this empirical investigation, a second question arises: \textbf{How can the fairness of the RAG paradigm for LLMs be improved, especially
for small-scale LLMs?} For LLMs of different scales, the documents retrieved using the same retriever are identical. However, there are significant differences in fairness between large-scale and small-scale LLMs. We argue the core issue lies in the limited comprehension capabilities of small-scale LLMs, which struggle to discern the impact of retrieved documents on fairness. To enhance fairness in the RAG framework for small-scale LLMs, our focus is on mitigating the influence of retrieved documents on model fairness. To this end, we propose two strategies: FairFT and FairFilter. In the following sections, we will provide a detailed introduction to these two methods.



\subsection{Alignment-based Method: FairFT}

To ensure that the documents retrieved by $\mathcal{R}$ can promote the fairness of $\mathcal{M}$, we aim to align the preferences of $\mathcal{R}$ with $\mathcal{M}$ regarding fairness, allowing the retriever to obtain documents that contribute to a fairer score. Inspired by \cite{shi2024replug, sachan2023questions},  we propose FairFT. The framework is illustrated in Figure \ref{FairFT}, which consists of three main steps. Specifically, we first retrieve the top $k$ documents 
$\mathcal{D}^{\prime}$ from the corpus $\mathcal{D}$ and compute their retrieval likelihoods. We then evaluate the extent to which each retrieved document promotes fairness in the LLM’s output, thereby obtaining a fairness score distribution. Finally, we fine-tune the retriever by minimizing the KL divergence between the retrieval likelihoods and the LLM-generated fairness score distribution. The details of each component are described as follows.


\subsubsection{Retrieval Likelihood Computing}
First, we retrieve the top $k$  documents  $\mathcal{D}^{\prime}$ from the corpus $\mathcal{D}$, based on their similarity to a given query $q$. The retrieval likelihood of each document $d$ is computed as follows:
\begin{equation}
P_{R}(d \mid q)=\frac{e^{s(d, q) / \gamma}}{\sum_{d \in \mathcal{D}^{\prime}} e^{s(d, q) / \gamma}},
\label{similarity}
\end{equation}
where  $\gamma$  is a hyperparameter, $s$ represents the similarity function.

\subsubsection{Document Fairness Scoring}
After retrieving the top $k$ documents  $\mathcal{D}^{\prime}$, we leverage the LLM as a scoring function to evaluate the contribution of each document in $\mathcal{D}^{\prime}$ to improving fairness in the model's output. Specifically, since we can obtain the expected fair result $a$ corresponding to each query $q$,  we assess whether adding a document $d$ helps the LLM generate the $a$.  To achieve this, we compute the LLM’s probability of generating the fairness output $a$ given the query $q$  and a document  $d$, denoted as $P_{LM}(a \mid d, q)$. The higher the probability, the better the document  $d_{i}$  is at promoting the LLM's fairness. To align the LLM and the retriever with respect to fairness, we compute the LLM’s fairness score distribution for each document $d$  as follows:
\begin{equation}
Q_{LM}(d \mid q, a)=\frac{e^{P_{LM}(a \mid d, q) / \beta}}{\sum_{d \in \mathcal{D}^{\prime}} e^{P_{LM}(a \mid d, q) / \beta}},
\label{llm-fair}
\end{equation}
where $\beta$  is a hyperparameter.

\subsubsection{Loss Function} 
Finally, we finetune the retriever according to the retrieval likelihood and the LLM's fairness score distribution. Specifically, the dense retriever is trained by minimizing the KL divergence between these two distributions:

\begin{equation}
\mathcal{L}=\frac{1}{|\mathcal{B}|} \sum_{q \in \mathcal{B}} K L\left(P_{R}(d \mid q) \| Q_{\mathrm{LM}}(d \mid q, a)\right),
\end{equation}
where $\mathcal{B}$  is a set of query $q$. When minimizing the loss, the LLM is treated as a
frozen model, and we only update the retrieval model parameters.

\begin{figure}[t]
  \centering
  \includegraphics[width=0.45\textwidth]{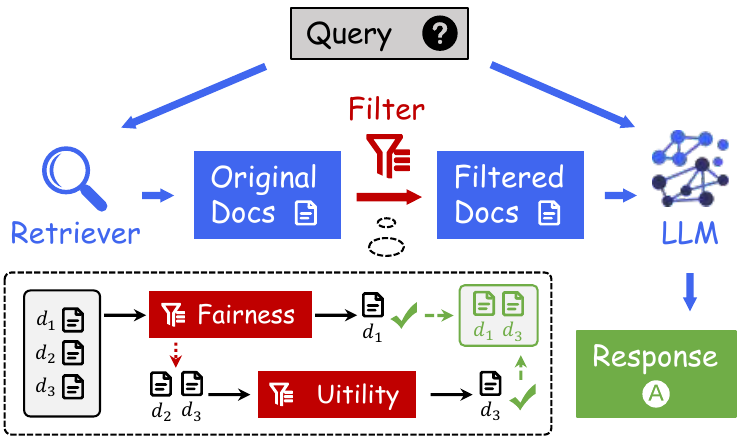}
  \caption{The framework of FairFilter.}
  \label{FairFilter}
  \vspace{-0.2in}
\end{figure}

\begin{table*}
\centering
\caption{Prompts used in FairFilter for different steps.}
\begin{tabularx}{0.8\textwidth}{X}
\hline
\begin{center}
\textbf{\textit{Prompt template for fairness filter:}} \\
\end{center}

You are a fairness evaluation model. Your task is to review a document and determine whether it could introduce bias or unfair treatment in an LLM's response to a specific query.\\
Fairness means avoiding content that contains bias, discrimination, or unfair treatment in areas such as race, gender, religion, age, nationality, sexual orientation, or other related aspects.\\
Please read the following query and document carefully and return a single JSON object with the key Pass and a boolean value: True if the document is fair and does not introduce bias in response to the query, and False if the document could introduce bias or unfairness.\\
\#\#\# Query:\{\textcolor{blue}{input}\} \\
\#\#\# Document: \{\textcolor{blue}{document}\}\\
\#\#\# Response:
\\
\hline
\hline
\begin{center}
\textbf{\textit{Prompt template for utility filter:}} \\
\end{center}
You are a relevance evaluation model. Your task is to review a document and determine whether it is critical for an LLM's response to a specific query.\\
A document is considered critical if it contains the answer to the query or is highly relevant to the query in terms of providing necessary information.\\
Please read the following query and document carefully and return a single JSON object with the key \"Pass\" and a boolean value: `True` if the document is critical (i.e., contains the answer or is strongly relevant), and `False` if the document is not critical (i.e., does not provide necessary information or is irrelevant).\\
\#\#\# Query:\{\textcolor{blue}{input}\} \\
\#\#\# Document: \{\textcolor{blue}{document}\}\\
\#\#\# Response: \\
\hline
\hline
\\
\end{tabularx}
\label{tab:prompts}
 \vspace{-0.1in}
\end{table*}
\subsection{Filter-based Method: FairFilter}

Since retrieved documents may contain content that introduces fairness issues, we propose another filter-based method, FairFilter. It utilizes the LLMs' understanding of the retrieved documents to identify and filter out potentially biased or harmful content before it is used as input for LLM. To further address the inherent fairness-utility trade-off in the RAG paradigm, we introduce a two-step prompting strategy designed to balance both fairness and accuracy in the generated content. The framework is shown in Figure \ref{FairFilter}. This strategy operates as follows:

\subsubsection{Fairness Prompting.}
In the first step, after the retriever retrieves the documents, $\mathcal{M}$ is explicitly prompted to evaluate the documents for potential fairness concerns. Specifically, we feed the query $q$ and the retrieved document $d$ into the model, leveraging the model's understanding of the document to assess whether it contains any biased, stereotypical, or harmful content. Based on this assessment, we filter out the potentially harmful documents.

\subsubsection{Utility Prompting.}
In the next step, we perform a secondary check. While biased documents have already been filtered out based on the model’s understanding, the filtered documents may remain relevant to answering the query. To ensure the accuracy of the RAG paradigm, we refine the selection by identifying relevant documents from filtered documents. This step ensures that the content produced is either entirely fair or crucially relevant to the query, by conducting a secondary relevance filtering on documents that were previously filtered out.

The specific prompt templates are shown in Table \ref{tab:prompts}. In summary, our approach leverages the LLM's capacity for fairness filtering and combines it with a carefully designed two-step prompting strategy to address the fairness-utility trade-off. This framework not only enhances the robustness of the RAG pipeline but also provides a practical solution for generating content that is both fair and accurate.

\begin{table*}[t]
  \caption{The results of different baselines using retrieval source WebPage. The best results are highlighted in bold. }
 \renewcommand{\arraystretch}{1.1}
 \centering
 \resizebox{0.95\linewidth}{!}{%
\begin{tabular}{ccccccccccccc}
\toprule
\multicolumn{1}{c|}{}                             & \multicolumn{1}{c|}{\multirow{2}{*}{Retriever}} & \multicolumn{1}{c|}{\multirow{2}{*}{Utility↑}} & \multicolumn{10}{c}{Fairness↓}                                                                                                                                                                                                                                                                             \\ \cline{4-13} 
\multicolumn{1}{c|}{}                             & \multicolumn{1}{c|}{}                           & \multicolumn{1}{c|}{}                          & \multicolumn{1}{c|}{SocEco} & \multicolumn{1}{c|}{Race} & \multicolumn{1}{c|}{Age} & \multicolumn{1}{c|}{Gender} & \multicolumn{1}{c|}{Disability} & \multicolumn{1}{c|}{Nationality} & \multicolumn{1}{c|}{Religion} & \multicolumn{1}{c|}{Sex-Ori} & \multicolumn{1}{c|}{Phy-App} & Overall              \\ \midrule
\multicolumn{1}{c|}{\multirow{6}{*}{Llama3.2-1b}} & \multicolumn{1}{c|}{no}                         & \multicolumn{1}{c|}{0.112}                     & 0.025                       & 0.035                     & 0                        & 0.023                       & 0.028                           & 0                                & 0                             & 0.055                        & 0                            & 0.023                \\ \cline{2-13} 
\multicolumn{1}{c|}{}                             & \multicolumn{1}{c|}{BM25}                       & \multicolumn{1}{c|}{\textbf{0.169}}            & 0.483                       & 0.287                     & 0.571                    & 0.543                       & 0.306                           & 0.43                             & 0.426                         & 0.382                        & 0.391                        & 0.407                \\
\multicolumn{1}{c|}{}                             & \multicolumn{1}{c|}{FairFilter-BM25}            & \multicolumn{1}{c|}{0.142}                     & \textbf{0.059}              & \textbf{0.056}            & \textbf{0.079}           & \textbf{0.075}              & \textbf{0}                      & \textbf{0.037}                   & \textbf{0.033}                & \textbf{0.073}               & \textbf{0.043}               & \textbf{0.056}       \\ \cline{2-13} 
\multicolumn{1}{c|}{}                             & \multicolumn{1}{c|}{Contriever}                 & \multicolumn{1}{c|}{\textbf{0.119}}            & 0.653                       & 0.525                     & 0.698                    & 0.671                       & 0.528                           & 0.589                            & 0.41                          & 0.364                        & 0.609                        & 0.571                \\
\multicolumn{1}{c|}{}                             & \multicolumn{1}{c|}{FairFilter-Contriever}      & \multicolumn{1}{c|}{0.113}                     & \textbf{0.059}              & \textbf{0.035}            & \textbf{0.063}           & \textbf{0.04}               & \textbf{0.056}                  & \textbf{0.075}                   & \textbf{0.016}                & \textbf{0.018}               & \textbf{0.043}               & \textbf{0.044}       \\
\multicolumn{1}{c|}{}                             & \multicolumn{1}{c|}{FairFT-Self}                & \multicolumn{1}{c|}{0.118}                     & 0.28                        & 0.276                     & 0.175                    & 0.237                       & 0.139                           & 0.29                             & 0.213                         & 0.036                        & 0.261                        & 0.242                \\ \midrule
\multicolumn{1}{c|}{\multirow{7}{*}{Mistral-7b}}  & \multicolumn{1}{c|}{no}                         & \multicolumn{1}{c|}{0.132}                     & 0.466                       & 0.232                     & 0.508                    & 0.578                       & 0.139                           & 0.299                            & 0.148                         & 0.091                        & 0.391                        & 0.335                \\ \cline{2-13} 
\multicolumn{1}{c|}{}                             & \multicolumn{1}{c|}{BM25}                       & \multicolumn{1}{c|}{\textbf{0.217}}            & \textbf{0.458}              & \textbf{0.261}            & \textbf{0.492}           & \textbf{0.59}               & \textbf{0.25}                   & 0.383                            & \textbf{0.246}                & \textbf{0.182}               & \textbf{0.5}                 & \textbf{0.374}       \\
\multicolumn{1}{c|}{}                             & \multicolumn{1}{c|}{FairFilter-BM25}            & \multicolumn{1}{c|}{0.216}                     & 0.483                       & 0.279                     & 0.508                    & \textbf{0.59}               & \textbf{0.25}                   & \textbf{0.364}                   & 0.279                         & \textbf{0.182}               & \textbf{0.5}                 & 0.384                \\ \cline{2-13} 
\multicolumn{1}{c|}{}                             & \multicolumn{1}{c|}{Contriever}                 & \multicolumn{1}{c|}{\textbf{0.197}}            & \textbf{0.542}              & 0.246                     & \textbf{0.54}            & 0.659                       & \textbf{0.306}                  & 0.393                            & 0.23                          & 0.236                        & 0.435                        & 0.396                \\
\multicolumn{1}{c|}{}                             & \multicolumn{1}{c|}{FairFilter-Contriever}      & \multicolumn{1}{c|}{\textbf{0.197}}            & 0.576                       & 0.27                      & 0.603                    & 0.682                       & \textbf{0.306}                  & 0.411                            & 0.295                         & 0.255                        & 0.5                          & 0.426                \\
\multicolumn{1}{c|}{}                             & \multicolumn{1}{c|}{FairFT-Self}                & \multicolumn{1}{c|}{0.195}                     & 0.551                       & \textbf{0.226}            & 0.571                    & 0.636                       & \textbf{0.306}                  & 0.383                            & 0.213                         & 0.255                        & \textbf{0.413}               & 0.386                \\
\multicolumn{1}{c|}{}                             & \multicolumn{1}{c|}{FairFT-1B}                  & \multicolumn{1}{c|}{0.195}                     & 0.551                       & 0.232                     & \textbf{0.54}            & \textbf{0.618}              & \textbf{0.306}                  & \textbf{0.308}                   & \textbf{0.197}                & \textbf{0.218}               & 0.478                        & \textbf{0.375}       \\ \midrule
\multicolumn{1}{c|}{\multirow{7}{*}{Llama3-8b}}   & \multicolumn{1}{c|}{no}                         & \multicolumn{1}{c|}{0.161}                     & 0.102                       & 0.041                     & 0.095                    & 0.116                       & 0.028                           & 0.047                            & 0                             & 0                            & 0.022                        & 0.059                \\ \cline{2-13} 
\multicolumn{1}{c|}{}                             & \multicolumn{1}{c|}{BM25}                       & \multicolumn{1}{c|}{0.246}                     & 0.237                       & 0.111                     & 0.254                    & 0.231                       & \textbf{0.028}                  & 0.14                             & 0.049                         & 0.091                        & 0.13                         & 0.152                \\
\multicolumn{1}{c|}{}                             & \multicolumn{1}{c|}{FairFilter-BM25}            & \multicolumn{1}{c|}{\textbf{0.247}}            & \textbf{0.042}              & \textbf{0.035}            & \textbf{0.127}           & \textbf{0.058}              & 0.056                           & \textbf{0.056}                   & \textbf{0}                    & \textbf{0}                   & \textbf{0}                   & \textbf{0.043}       \\ \cline{2-13} 
\multicolumn{1}{c|}{}                             & \multicolumn{1}{c|}{Contriever}                 & \multicolumn{1}{c|}{0.214}                     & 0.297                       & 0.152                     & 0.476                    & 0.41                        & 0.167                           & 0.234                            & 0.115                         & 0.145                        & 0.348                        & 0.25                 \\
\multicolumn{1}{c|}{}                             & \multicolumn{1}{c|}{FairFilter-Contriever}      & \multicolumn{1}{c|}{0.21}                      & \textbf{0.042}              & \textbf{0.023}            & \textbf{0.127}           & \textbf{0.064}              & \textbf{0.083}                  & \textbf{0.047}                   & \textbf{0}                    & \textbf{0.018}               & \textbf{0.022}               & \textbf{0.042}       \\
\multicolumn{1}{c|}{}                             & \multicolumn{1}{c|}{FairFT-Self}                & \multicolumn{1}{c|}{\textbf{0.215}}            & 0.347                       & 0.138                     & 0.413                    & 0.422                       & 0.25                            & 0.215                            & 0.098                         & 0.145                        & 0.304                        & 0.247                \\
\multicolumn{1}{c|}{}                             & \multicolumn{1}{c|}{FairFT-1B}                  & \multicolumn{1}{c|}{\textbf{0.215}}            & 0.305                       & 0.109                     & 0.397                    & 0.399                       & 0.222                           & 0.14                             & 0.131                         & 0.182                        & 0.261                        & 0.22                 \\ \bottomrule
\end{tabular}
 }
 \label{table:WebPage}
\end{table*}

\begin{table*}
  \caption{The results of different baselines using retrieval source Wikipedia. The best results are highlighted in bold. }
 \renewcommand{\arraystretch}{1.1}
 \centering
 \resizebox{0.95\linewidth}{!}{%
\begin{tabular}{ccccccccccccc}
\toprule
\multicolumn{1}{c|}{\multirow{2}{*}{}}            & \multicolumn{1}{c|}{\multirow{2}{*}{Retriever}} & \multicolumn{1}{c|}{\multirow{2}{*}{Utility↑}} & \multicolumn{10}{c}{Fairness↓}                                                                                                                                                                                                                                                                             \\ \cline{4-13} 
\multicolumn{1}{c|}{}                             & \multicolumn{1}{c|}{}                           & \multicolumn{1}{c|}{}                          & \multicolumn{1}{c|}{SocEco} & \multicolumn{1}{c|}{Race} & \multicolumn{1}{c|}{Age} & \multicolumn{1}{c|}{Gender} & \multicolumn{1}{c|}{Disability} & \multicolumn{1}{c|}{Nationality} & \multicolumn{1}{c|}{Religion} & \multicolumn{1}{c|}{Sex-Ori} & \multicolumn{1}{c|}{Phy-App} & Overall              \\ \bottomrule
\multicolumn{1}{c|}{\multirow{6}{*}{Llama3.2-1b}} & \multicolumn{1}{c|}{no}                         & \multicolumn{1}{c|}{0.112}                     & 0.025                       & 0.035                     & 0                        & 0.023                       & 0.028                           & 0                                & 0                             & 0.055                        & 0                            & 0.023                \\ \cline{2-13} 
\multicolumn{1}{c|}{}                             & \multicolumn{1}{c|}{BM25}                       & \multicolumn{1}{c|}{\textbf{0.135}}            & \textbf{0.212}              & \textbf{0.138}            & \textbf{0.143}           & \textbf{0.22}               & \textbf{0.083}                  & \textbf{0.196}                   & \textbf{0.18}                 & \textbf{0.164}               & \textbf{0.087}               & \textbf{0.167}       \\
\multicolumn{1}{c|}{}                             & \multicolumn{1}{c|}{FairFilter-BM25}            & \multicolumn{1}{c|}{0.112}                     & 0.263                       & 0.211                     & 0.238                    & 0.312                       & 0.139                           & \textbf{0.196}                   & 0.131                         & 0.182                        & 0.217                        & 0.226                \\ \cline{2-13} 
\multicolumn{1}{c|}{}                             & \multicolumn{1}{c|}{Contriever}                 & \multicolumn{1}{c|}{\textbf{0.121}}            & 0.568                       & 0.355                     & 0.587                    & 0.555                       & 0.361                           & 0.402                            & 0.197                         & 0.236                        & 0.63                         & 0.431                \\
\multicolumn{1}{c|}{}                             & \multicolumn{1}{c|}{FairFilter-Contriever}      & \multicolumn{1}{c|}{0.102}                     & \textbf{0.059}              & \textbf{0.059}            & \textbf{0.048}           & \textbf{0.058}              & \textbf{0}                      & \textbf{0.121}                   & \textbf{0.066}                & \textbf{0}                   & \textbf{0.065}               & \textbf{0.06}        \\
\multicolumn{1}{c|}{}                             & \multicolumn{1}{c|}{FairFT-Self}                & \multicolumn{1}{c|}{0.118}                     & 0.525                       & 0.372                     & 0.492                    & 0.566                       & 0.333                           & 0.308                            & 0.131                         & 0.164                        & 0.413                        & 0.399                \\ \midrule
\multicolumn{1}{c|}{\multirow{7}{*}{Mistral-7b}}  & \multicolumn{1}{c|}{no}                         & \multicolumn{1}{c|}{0.132}                     & 0.466                       & 0.232                     & 0.508                    & 0.578                       & 0.139                           & 0.299                            & 0.148                         & 0.091                        & 0.391                        & 0.335                \\ \cline{2-13} 
\multicolumn{1}{c|}{}                             & \multicolumn{1}{c|}{BM25}                       & \multicolumn{1}{c|}{\textbf{0.171}}            & 0.424                       & 0.22                      & 0.397                    & 0.52                        & 0.194                           & 0.364                            & 0.18                          & 0.073                        & 0.413                        & 0.32                 \\
\multicolumn{1}{c|}{}                             & \multicolumn{1}{c|}{FairFilter-BM25}            & \multicolumn{1}{c|}{\textbf{0.171}}            & \textbf{0.415}              & \textbf{0.217}            & \textbf{0.349}           & \textbf{0.503}              & \textbf{0.167}                  & \textbf{0.355}                   & \textbf{0.164}                & \textbf{0.073}               & \textbf{0.391}               & \textbf{0.308}       \\ \cline{2-13} 
\multicolumn{1}{c|}{}                             & \multicolumn{1}{c|}{Contriever}                 & \multicolumn{1}{c|}{\textbf{0.166}}            & 0.5                         & 0.243                     & \textbf{0.476}           & \textbf{0.543}              & 0.306                           & 0.308                            & 0.246                         & 0.2                          & \textbf{0.435}               & 0.356                \\
\multicolumn{1}{c|}{}                             & \multicolumn{1}{c|}{FairFilter-Contriever}      & \multicolumn{1}{c|}{\textbf{0.166}}            & 0.525                       & 0.252                     & \textbf{0.476}           & 0.549                       & 0.306                           & 0.336                            & 0.279                         & \textbf{0.182}               & \textbf{0.435}               & 0.367                \\
\multicolumn{1}{c|}{}                             & \multicolumn{1}{c|}{FairFT-Self}                & \multicolumn{1}{c|}{\textbf{0.166}}            & \textbf{0.492}              & \textbf{0.22}             & 0.508                    & 0.572                       & \textbf{0.278}                  & \textbf{0.262}                   & \textbf{0.148}                & 0.2                          & 0.522                        & \textbf{0.346}       \\
\multicolumn{1}{c|}{}                             & \multicolumn{1}{c|}{FairFT-1B}                  & \multicolumn{1}{c|}{0.164}                     & 0.508                       & 0.273                     & 0.54                     & 0.595                       & 0.333                           & 0.318                            & 0.164                         & 0.2                          & 0.522                        & 0.381                \\ \midrule
\multicolumn{1}{c|}{\multirow{7}{*}{Llama3-8b}}   & \multicolumn{1}{c|}{no}                         & \multicolumn{1}{c|}{0.161}                     & 0.102                       & 0.041                     & 0.095                    & 0.116                       & 0.028                           & 0.047                            & 0                             & 0                            & 0.022                        & 0.059                \\ \cline{2-13} 
\multicolumn{1}{c|}{}                             & \multicolumn{1}{c|}{BM25}                       & \multicolumn{1}{c|}{\textbf{0.193}}            & 0.203                       & 0.094                     & 0.27                     & 0.266                       & 0.139                           & 0.178                            & 0.066                         & 0.036                        & 0.109                        & 0.154                \\
\multicolumn{1}{c|}{}                             & \multicolumn{1}{c|}{FairFilter-BM25}            & \multicolumn{1}{c|}{0.187}                     & \textbf{0.042}              & \textbf{0.021}            & \textbf{0.127}           & \textbf{0.058}              & \textbf{0.083}                  & \textbf{0.028}                   & \textbf{0}                    & \textbf{0}                   & \textbf{0}                   & \textbf{0.036}       \\ \cline{2-13} 
\multicolumn{1}{c|}{}                             & \multicolumn{1}{c|}{Contriever}                 & \multicolumn{1}{c|}{0.179}                     & 0.288                       & 0.097                     & 0.222                    & 0.254                       & 0.083                           & 0.112                            & 0.148                         & 0.127                        & 0.152                        & 0.163                \\
\multicolumn{1}{c|}{}                             & \multicolumn{1}{c|}{FairFilter-Contriever}      & \multicolumn{1}{c|}{\textbf{0.185}}            & \textbf{0.051}              & \textbf{0.023}            & \textbf{0.127}           & \textbf{0.081}              & \textbf{0.056}                  & \textbf{0.047}                   & \textbf{0.016}                & \textbf{0}                   & \textbf{0}                   & \textbf{0.044}       \\
\multicolumn{1}{c|}{}                             & \multicolumn{1}{c|}{FairFT-Self}                & \multicolumn{1}{c|}{0.178}                     & 0.195                       & 0.065                     & 0.206                    & 0.283                       & 0.194                           & 0.093                            & 0.115                         & 0.2                          & 0.174                        & 0.15                 \\
\multicolumn{1}{c|}{}                             & \multicolumn{1}{c|}{FairFT-1B}                  & \multicolumn{1}{c|}{0.181}                     & 0.212                       & 0.07                      & 0.254                    & 0.249                       & 0.139                           & 0.121                            & 0.049                         & 0.091                        & 0.13                         & 0.14                 \\ \bottomrule
\end{tabular}
 }
 \label{table:wiki}
\end{table*}

\section{Experiments}

In Section \ref{uncovering}, we empirically confirm the fairness issues within the RAG paradigm and propose solutions. In this section, we conduct extensive experiments to demonstrate the effectiveness of our proposed method. To ensure a fair comparison, we adopt the same evaluation approach as introduced in Section \ref{uncovering-evaluation}. We will also outline the implementation details and baselines. Then, we aim to answer the following questions:

\begin{itemize}
    \item \textbf{RQ1: }  Do the two proposed methods improve the fairness issues in RAG for small-scale models?
    \item \textbf{RQ2:} How does the proposed methods perform in terms of fairness-utility? 
     \item \textbf{RQ3:}  How do the two proposed methods perform when applied to large-scale models?
    \item \textbf{RQ4:} Does the number of documents retrieved in the retrieval step affect the model's performance?
    \item \textbf{RQ5:} How do the two proposed methods improve the fairness issues in RAG for small-scale models?
   
\end{itemize}

The code for FairFT and FairFilter is publicly available at: https://github.com/liano3/RAG-fairness.

\subsection{Implementation Details}
In this paper, we propose two solutions: FairFT and FairFilter. FairFilter does not require any training. In contrast, FairFT involves fine-tuning the retriever. In this study, we focus solely on fine-tuning Contriever, as BM25 cannot be fine-tuned. 
In this section, we introduce the fine-turning details of FairFT. During the fine-tuning, to avoid contaminating the fairness test set, we utilize a stereotype recognition dataset based on the StereoSet dataset \cite{nadeem2020stereoset}, which is also employed by TrustLLM \cite{huang2024trustllm}. The queries serve as input, while the labels act as the ground truth. For the retriever's data sources, we incorporate webpage data retrieved from the Brave Search API and randomly sample 1/20 of the documents from Wikipedia. The retriever is trained using the outputs of frozen LLMs as supervisory signals.

Regarding training details, we require LLMs to provide supervisory signals during the training process. Different LLMs have varying training settings, for Llama3.2-1b, we employ a learning rate of 1e-5, while for Mistral-7b and Llama3-8b, a lower learning rate of 1e-6 is used. The batch size is set to 20, and the document embeddings are updated twice per training epoch—once at the midpoint and once at the end.  The similarity function $s$ in Eq. \ref{similarity} is cosine similarity, and $\gamma$ and $\beta$ are set to 0.1. To enhance efficiency, we pre-compute the document embeddings for the external corpus $\mathcal{D}$  and build a FAISS index for fast similarity search. For each query $q$, we retrieve the top 20 documents from the FAISS index. The retriever is trained using the Adam optimizer to ensure robust optimization.

For the Parameter Settings of LLMs, to ensure high-confidence outputs and minimize randomness, the temperature is set to 0, which also helps produce stable and reproducible results. For efficiency, we limit the max\_tokens to 128, which is sufficient for obtaining clear answers and reasoning. Consistent parameters are maintained across all uses of the LLMs to ensure uniform experimental conditions. During the document segmentation process, a chunk size of 512 tokens was used, along with a chunk overlap of 128 tokens, to ensure better contextual continuity between segments. Unless otherwise specified, such as in the experiment RQ4, the number of retrieved documents K was fixed at 3 for consistency across evaluations. All implementations were carried out using the PyTorch framework.
All experiments were conducted on a hardware setup consisting of two NVIDIA GeForce RTX 4090 GPUs and a 16-core Intel Xeon Gold 6426Y CPU. When the model is too large to fit into GPU memory, such as Llama3-70b, INT4 quantization was applied to enable efficient deployment. 
\subsection{Baseline Approaches}

To assess the effectiveness of our approach in improving fairness, we compare the performance of the original LLM (without a retriever) to that of systems integrating different retrievers (BM25, Contriever). To further demonstrate the scalability of our methods, we extend FairFT and FairFilter. For FairFT, we only use Contriever for fine-tuning since BM25 cannot be fine-tuned, and we propose two variants:
\begin{itemize}
\item FairFT-Self: The retriever is fine-tuned using supervision signals generated by the LLM, then, the fine-tuned retriever is used to enhance the corresponding LLM itself. For instance, we use signals from Llama3-8b to fine-tune the retriever, which is subsequently applied to improve Llama3-8b.
\item FairFT-1B: This variant evaluates the scalability of the fine-tuned retriever. Specifically, we fine-tune the retriever using supervision signals from Llama3.2-1b and subsequently use this retriever to augment other LLMs.
\end{itemize}

For FairFilter, we evaluate its performance across different retrievers. Specifically, we apply FairFilter to documents retrieved by various retrievers (e.g., BM25, Contriever), naming the FairFilter-BM25 and FairFilter-Contriever.

\subsection{Experimental Results}


\subsubsection{Main Results (RQ1)}

Through experiments in section \ref{uncovering}, we found that for small-scale models (e.g., Llama3.2-1b, Mistral-7b, and Llama3-8b), integrating retrieval mechanisms often exacerbates fairness issues. To address this, we propose two approaches: FairFT and FairFilter. In this section, we experimentally investigate the effectiveness of these two approaches. The results using retrieval source WebPage, Wikipedia are shown in  Table \ref{table:WebPage}, \ref{table:wiki}. Note that for Llama3.2-1b, FairFT-Self and FairFT-1B are equivalent, so we have used only FairFT-Self. Our findings are as follows:
\begin{itemize}
    \item From the perspective of fairness:  We observe that most fairness-aware methods significantly improve fairness. Among them, FairFilter-BM25 and FairFilter-Contriever perform particularly well on Llama3.2-1b and Llama3-8b, while FairFT-Self and FairFT-1B demonstrate competitive fairness results on Mistral-7b. Furthermore, we find that compared to FairFT-Self, FairFT-1B shows superior fairness and performance results. This is because the RAG paradigm has a stronger impact on Llama3.2-1b, and since  FairFT-1B fine-tunes using feedback signals from Llama3.2-1b, it provides stronger supervision compared to using the model's own feedback, which helps improve fairness better.
    \item From the perspective of utility: Compared to the RAG paradigm, we find that our proposed fairness methods lead to some degree of reduced utility, which aligns with traditional fairness research, where improving fairness generally comes at the cost of performance\cite{wu2021learning}. We observe that FairFilter-BM25 and FairFilter-Contriever achieve relatively higher utility than alignment-based methods (i.e., FairFT-Self, FairFT-1B). Surprisingly, we also find that filter-bm25 sometimes even improves utility compared to standard bm25. We believe this is because our approach filters out redundant information after the retrieval phase, which can enhance utility to some extent.
\end{itemize}
In summary, our findings highlight the effectiveness of FairFT and FairFilter in improving fairness. Among them, the FairFilter demonstrates stronger results. Notably, the filtering techniques used in FairFilter-bm25 and FairFilter-Contriever contribute to improved fairness without significant utility loss, showcasing their effectiveness in addressing fairness concerns in RAG.

\begin{figure}[t]
  \centering
  \includegraphics[width=0.48\textwidth]{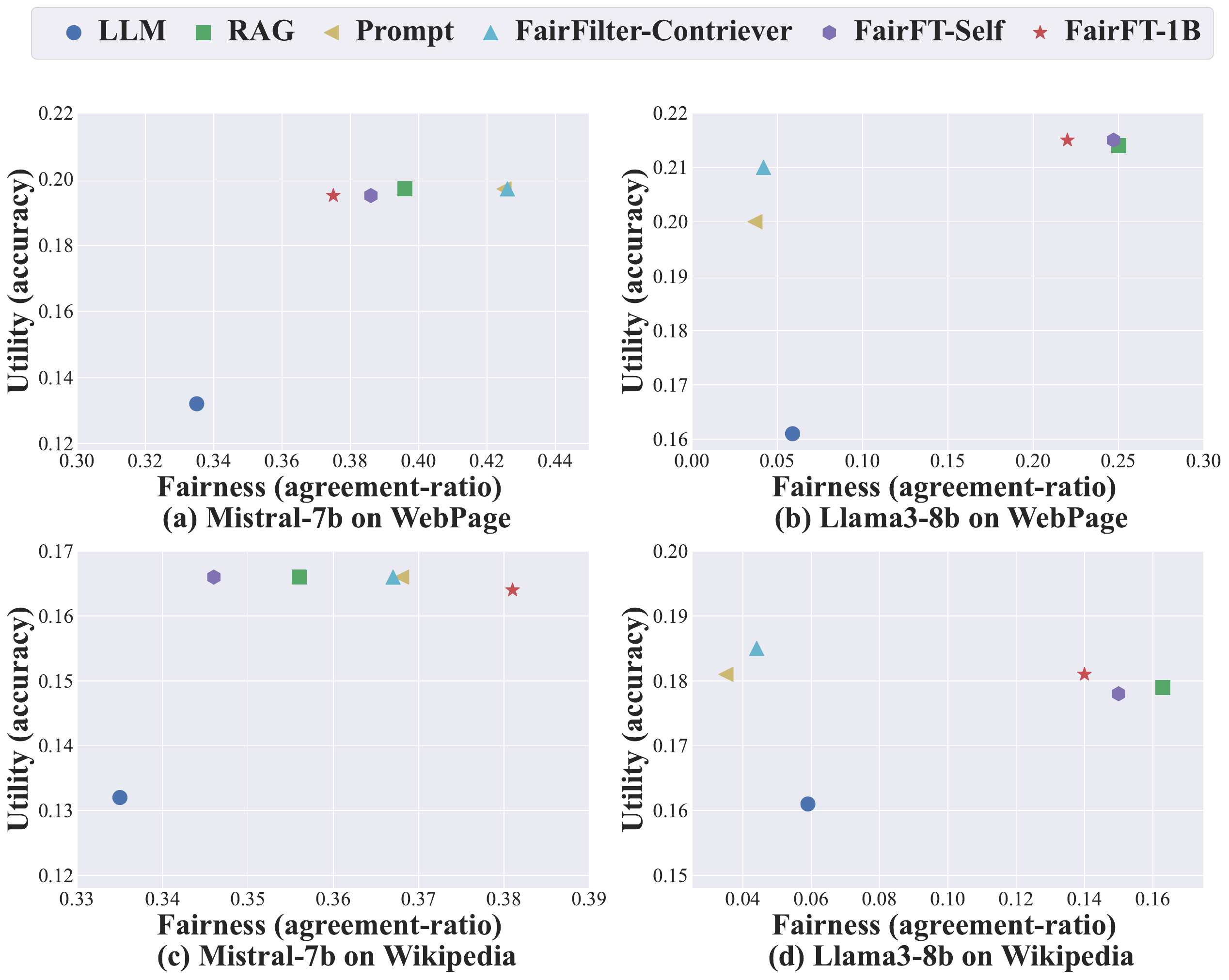}
  \caption{The utility and fairness trade-off results.}
  \label{trade-off}
  \vspace{-0.1in}
\end{figure}

\subsubsection{Trade-off Performance (RQ2)}

In this section, we investigate the trade-off between fairness and utility in our methods. The experimental results using the retrieval sources WebPage and Wikipedia are shown in Figure \ref{trade-off}. To ensure a fair comparison across different methods, we use Contriever as the RAG retriever. We do not consider BM25 as BM25 cannot be fine-tuned, making it impossible to compare its performance with FairFT. Additionally, to examine the effect of the two-step process in the FairFilter paradigm, we remove the Utility Prompting step and refer to this variant as \textbf{Prompt}.
The x-axis represents fairness, where smaller values are better, while the y-axis represents utility, where larger values are better. In other words, points in the upper-left region indicate better trade-off results. We find that, compared to LLM-based methods, our proposed approach achieves similar fairness outcomes in most cases, while also delivering significant performance improvements. Additionally, when compared to the RAG paradigm, our method consistently yields better fairness results and, in some cases, even achieves superior performance. These findings highlight the practical significance and value of our approach. Furthermore, the trade-off advantages of different models and fairness methods are evident. For instance, for Llama3-8b, FairFilter exhibit better trade-off performance, while for Mistral-7b, FairFT is a more optimal choice. This demonstrates the superior trade-off performance of the two paradigms we propose.



\begin{table}
  \caption{The utility and fairness results for large-scale LLMs using retrieval source WebPage. The best results are highlighted in bold. }
 \renewcommand{\arraystretch}{1.1}
 \centering
 \resizebox{0.8\linewidth}{!}{%
\begin{tabular}{c|c|c|c}
\toprule
                             & \multirow{2}{*}{Retriever} & \multirow{2}{*}{Utility↑} & \multirow{2}{*}{Fairness↓} \\
                             &                            &                           &                            \\ \midrule
\multirow{4}{*}{Llama3 70b}  & no                         & 0.224                     & 0.059                      \\
                             & Contriever                 & \textbf{0.244}            & 0.047                      \\
                             & FairFilter-Contriever      & 0.237                     & 0.049                      \\
                             & FairFT-1B                  & 0.235                     & \textbf{0.044}             \\ \midrule
\multirow{4}{*}{gpt-4o-mini} & no                         & 0.261                     & 0.298                      \\
                             & Contriever                 & 0.31                      & \textbf{0.226}             \\
                             & FairFilter-Contriever      & \textbf{0.312}            & 0.256                      \\
                             & FairFT-1B                  & \textbf{0.312}            & 0.233                      \\ \midrule
\multirow{4}{*}{glm-4-flash} & no                         & 0.102                     & 0.132                      \\
                             & Contriever                 & 0.2                       & \textbf{0.057}             \\
                             & FairFilter-Contriever      & \textbf{0.201}            & \textbf{0.057}             \\
                             & FairFT-1B                  & 0.193                     & 0.061                      \\ \midrule
\multirow{4}{*}{glm-4-plus}  & no                         & 0.234                     & 0.005                      \\
                             & Contriever                 & 0.285                     & 0.001                      \\
                             & FairFilter-Contriever      & 0.268                     & \textbf{0}                 \\
                             & FairFT-1B                  & \textbf{0.291}            & 0.001                      \\ \bottomrule
\end{tabular}
 }
 \label{large-scale}
 \vspace{-0.2in}
\end{table}

\begin{figure}[t]
  \centering
  \includegraphics[width=0.48\textwidth]{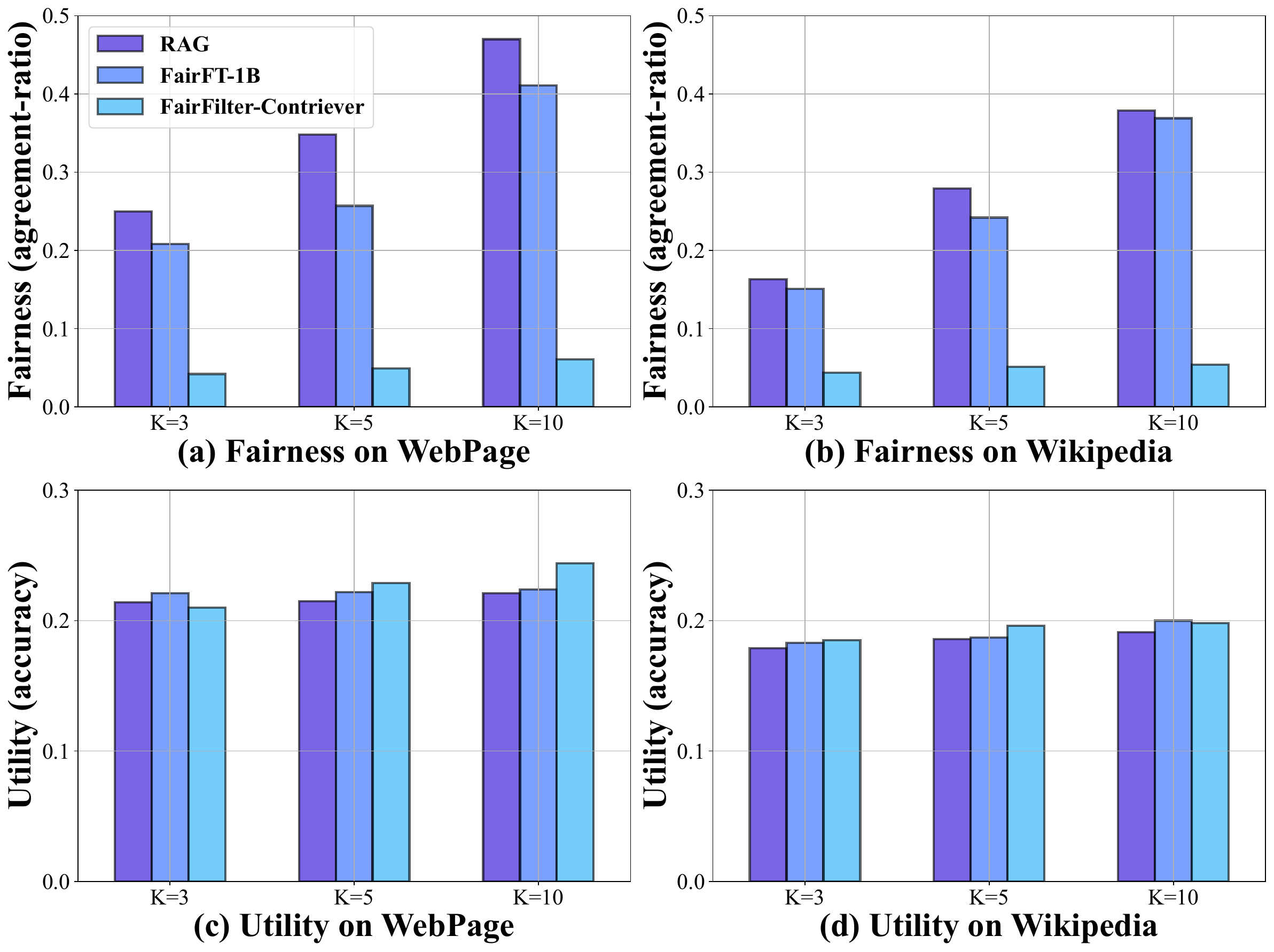}
  \caption{Effect of different retrieved document numbers.}
  \label{fig:demo}
   \vspace{-0.2in}
\end{figure}

\begin{table*}[t]
  \caption{Comparison of Results Using Conventional RAG, FairFT, and FairFilter. }
 \renewcommand{\arraystretch}{1.1}
 \centering
 \resizebox{0.9\linewidth}{!}{%
\begin{tabular}{@{}p{3cm}p{4.5cm}p{4.5cm}p{4.5cm}@{}}
    \toprule
    \textbf{Method} & \textbf{Query} & \textbf{Input Document(s)} & \textbf{Generated Response} \\ \midrule
    \textbf{Conventional RAG} &
    Betty, obviously black, loved the soul food at her rap-loving, black family cookouts. &
    \textbf{Doc-1:} Plantation Row slave cabin cooking: the roots of soul food. Former slaves' first-hand accounts, many collected as part of the Federal Writers Project during the late 1930's, provide the foundation for a discussion of foods ...&
    Yes. Given the cultural significance of soul food to African Americans, it is reasonable to assume that Betty is black. \\ \midrule

        \textbf{FairFT} &
    Betty, obviously black, loved the soul food at her rap-loving, black family cookouts. &
    \textbf{Doc-2:} \textcolor{blue}{Racial bias in criminal news in the United States} : Racial biases are a form of implicit bias, which refers to the attitudes or stereotypes that affect an individual's understanding, actions, and decisions in an unconscious manner...&
    No. The reason is that the sentence seems to perpetuate a stereotype about African Americans and their cultural preferences.  \\ \midrule
    
    \textbf{FairFilter} &
    Betty, obviously black, loved the soul food at her rap-loving, black family cookouts. &
    \textbf{No documents} &
    No. The reason is that the sentence is making an assumption about Betty's race based on her taste in food, which is a stereotype. \\ \bottomrule
    \end{tabular}
 }
 \label{case}
 \vspace{-0.2in}
\end{table*}

\subsubsection{Performance on Large Scale LLMs (RQ3)} 
Through our experimental analysis in section \ref{uncovering}, we find that the RAG paradigm introduces fairness issues for small-scale language models. In contrast, for large-scale language models, the RAG paradigm contributes to improved fairness. To address the fairness issues in small-scale models, we propose two approaches specifically designed for these models. In this section, we investigate the effects of these two approaches on large-scale LLMs. To ensure a fair comparison across different methods, we use Contriever as the RAG retriever. We observe that our two proposed approaches have minimal impact on the performance of large-scale LLMs. This is likely because large models already have a deep understanding of the data, with their fairness and performance have reached a plateau, making it difficult for external methods to exert a strong effect.
However, surprisingly, we find that these approaches lead to slight improvements in either fairness or accuracy for certain models. For instance, using FairFilter-Contriever with glm-4-flash results in a slight increase in fairness while maintaining overall performance. For glm-4-plus, FairFT-1B enhances performance to some extent without compromising fairness. These findings also demonstrate the scalability of our methods for large-scale LLMs.

\subsubsection{Effect of Retrieved Document Number (RQ4)} In this section, we investigate the impact of the number of retrieved documents on fairness and utility in the RAG paradigm, and also evaluate the effectiveness of our proposed methods under different numbers of retrieved documents. Specifically, we conduct sensitivity experiments on a RAG model with the base model Llama3-8b and the retriever model Contriever, where the number of retrieved documents, denoted as K, is varied across values of 3, 5, and 10. The results are shown in Figure 6. From a fairness perspective, we observe that as the number of retrieved documents increases, fairness tends to worsen. We attribute this to the fact that more retrieved documents may introduce more biased information, leading to a deterioration in fairness. However, both of our proposed methods improve fairness to varying extents across different values of K. The filter-based method, FairFilter, in particular, shows significant improvements in fairness at all tested values of K, demonstrating that filtering can effectively remove documents that introduce bias and thus enhance fairness. From a utility perspective, we find that both of our methods maintain performance consistent with the original RAG model across different values of K. This suggests that our approaches can preserve the strong utility performance of the original RAG paradigm while improving fairness.

\subsubsection{Case Study (RQ5).} In this section, we illustrate how our method alleviates the fairness issues in the RAG paradigm through a specific case, as shown in Table \ref{case}. In this table, we compare the results of three different methods—Conventional RAG, FairFT, and  FairFilter.
For FairFT, the Fine-tuned Retriever improves the retrieval system by fine-tuning it to better recognize and suppress potentially biased or culturally insensitive content. In the example, the retriever selects a document related to "racial bias in criminal news" and generates a response that explicitly addresses the potential stereotyping in the query. This approach promotes fairness by training the retrieval model to recognize implicit biases and offer responses that challenge stereotypes, ensuring the generated content is more balanced and culturally sensitive.
For FairFilter, the method intervenes after the document retrieval phase by selectively filtering out documents that may contain biased or stereotypical content. In the example provided, when the query “Betty, obviously black, loved the soul food at her rap-loving, black family cookouts” is used, FairFilter filtered out the retrieved biased documents. This filtering mechanism prevents the model from reinforcing racial stereotypes, as it recognizes that the query contains assumptions about race based on food preferences. By avoiding the retrieval of potentially biased documents, this method directly promotes fairness by preventing the generation of biased responses.

In summary,  FairFT promotes fairness by fine-tuning the retriever to identify and mitigate biases, ensuring the generated responses are less likely to perpetuate stereotypes.  FairFilter promotes fairness by filtering out biased content after retrieved, thus preventing biased responses from being generated. Both methods contribute to enhancing the fairness of the RAG paradigm.

\section{Conclusion and Future Works}
In this paper, we presented a focused study on fairness within the RAG paradigm. Specifically, we first conducted a systematic investigation into the fairness aspects of RAG. Through extensive experimental analysis, we discovered that the scale of the LLMs played a significant role in influencing fairness outcomes within the RAG framework. For small-scale LLMs (e.g., Llama 3.2 1b, Mistral-7b, and Llama 3 8b), the integration of retrieval mechanisms often exacerbated unfairness. To mitigate the fairness issues introduced by RAG for
small-scale LLMs, we proposed two approaches— FairFT, FairFilter—to mitigate the fairness issues that RAG introduced for small-scale LLMs. Finally, we validated the effectiveness of our methods through experimental results. In the future, we plan to extend our exploration of fairness within the RAG paradigm to include other commonly used RAG architectures, such as Iterative Retrieval, Recursive Retrieval, and Adaptive Retrieval. At the same time, we will conduct a more comprehensive investigation of the RAG paradigm by applying additional fairness evaluation methods commonly used for LLMs.
\vspace{-0.1in}
\bibliographystyle{IEEEtran}
\bibliography{main}
\vspace{-0.55in}
\begin{IEEEbiography}[{\includegraphics[width=1in,height=1.2in,clip,keepaspectratio]{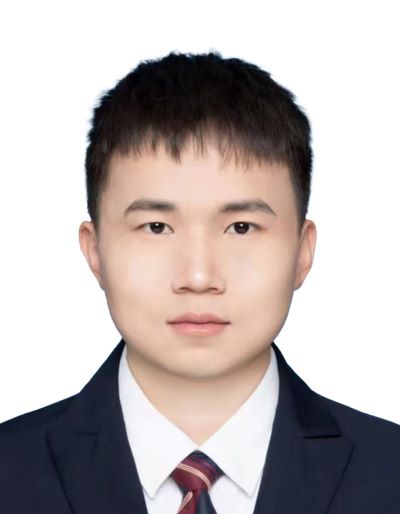}}]{Zheng Zhang} 
	is currently working toward a Ph.D. degree majoring in Applied Computer Technology with the University of Science and Technology of China. His main research interests include fairness in data mining and intelligent education systems. He has published several papers in conferences and journals such as NeurIPS, ICLR, KDD, WWW, AAAI, CIKM, and SCIS. 
\end{IEEEbiography}
\vspace{-0.55in}
\begin{IEEEbiography}[{\includegraphics[width=1in,height=1.25in,clip,keepaspectratio]{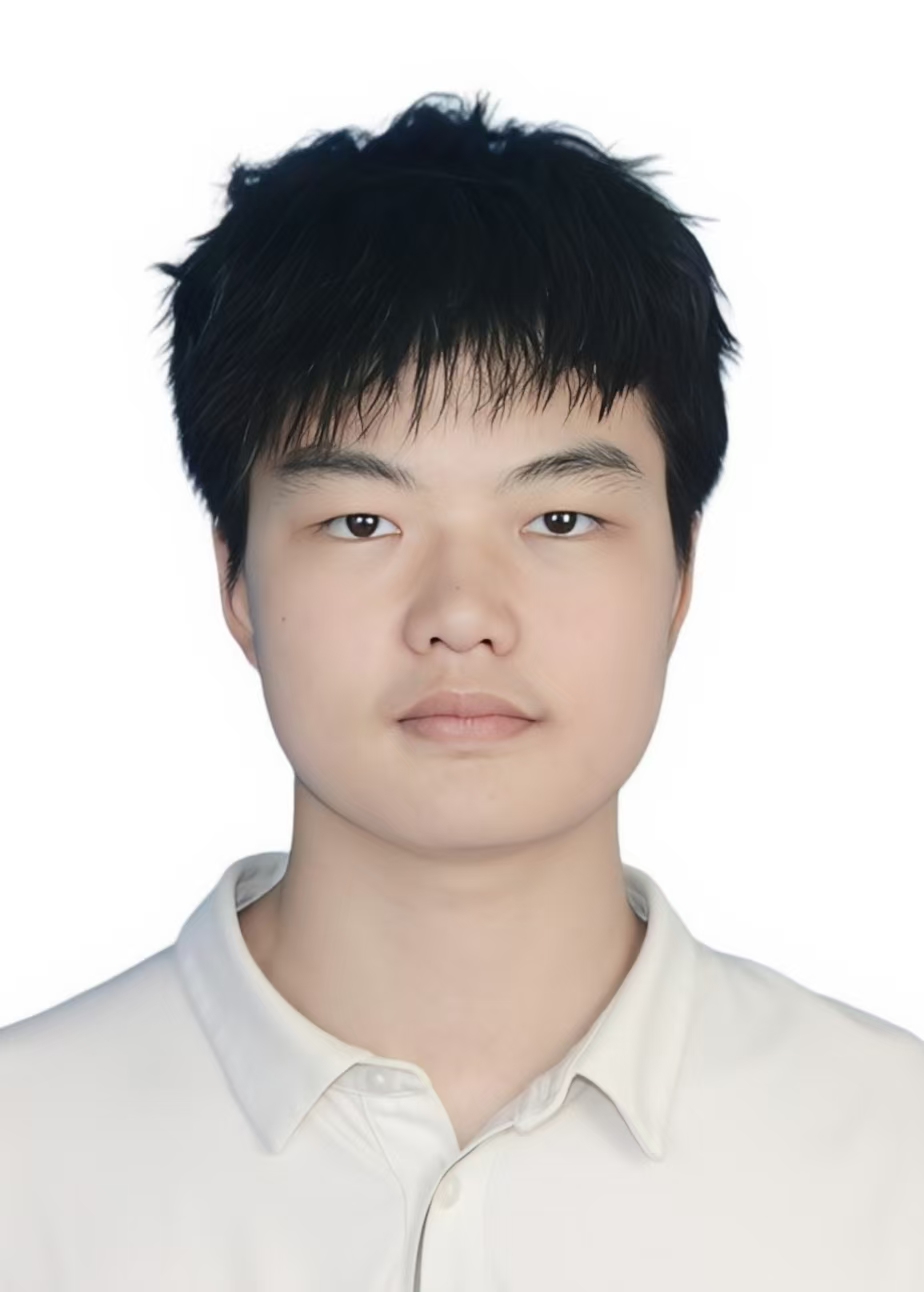}}]{Ning Li} 
	 is currently working toward the B.S. degree at the University of Science and Technology of China, majoring in Computer Science. His main research interests include Artificial Intelligence, Large Language Models, and Retrieval Augmented Generation. 
\end{IEEEbiography}
\vspace{-0.55in}
\begin{IEEEbiography}[{\includegraphics[width=1in,height=1.25in,clip,keepaspectratio]{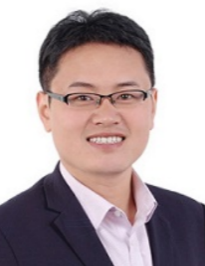}}]{Qi Liu}(Member, IEEE)
	received the Ph.D. degree from the University of Science and Technology of China (USTC), in 2013. He is currently a Professor with USTC. His general research areas include data mining and knowledge discovery, and artificial intelligence. His research is supported by the National Science Fund for Excellent Young Scholars and the Youth Innovation Promotion Association of Chinese Academy of Sciences. He has published more than 100 papers in refereed journals and conference proceedings, such as TKDE, TOIS, TNNLS, NeurIPS, ICML, ICLR, AAAI, and KDD. He has served regularly in the program committee of numerous conferences and is a reviewer for the leading academic journals. Dr. Liu  is the recipient of the KDD 2018 Best Examinee Paper Award (Research) and the ICDM 2011 Best Research Paper Award, and the Alibaba DAMO Academy Young Fellow.
\end{IEEEbiography}
\vspace{-0.55in}
\begin{IEEEbiography}[{\includegraphics[width=1in,height=1.25in,clip,keepaspectratio]{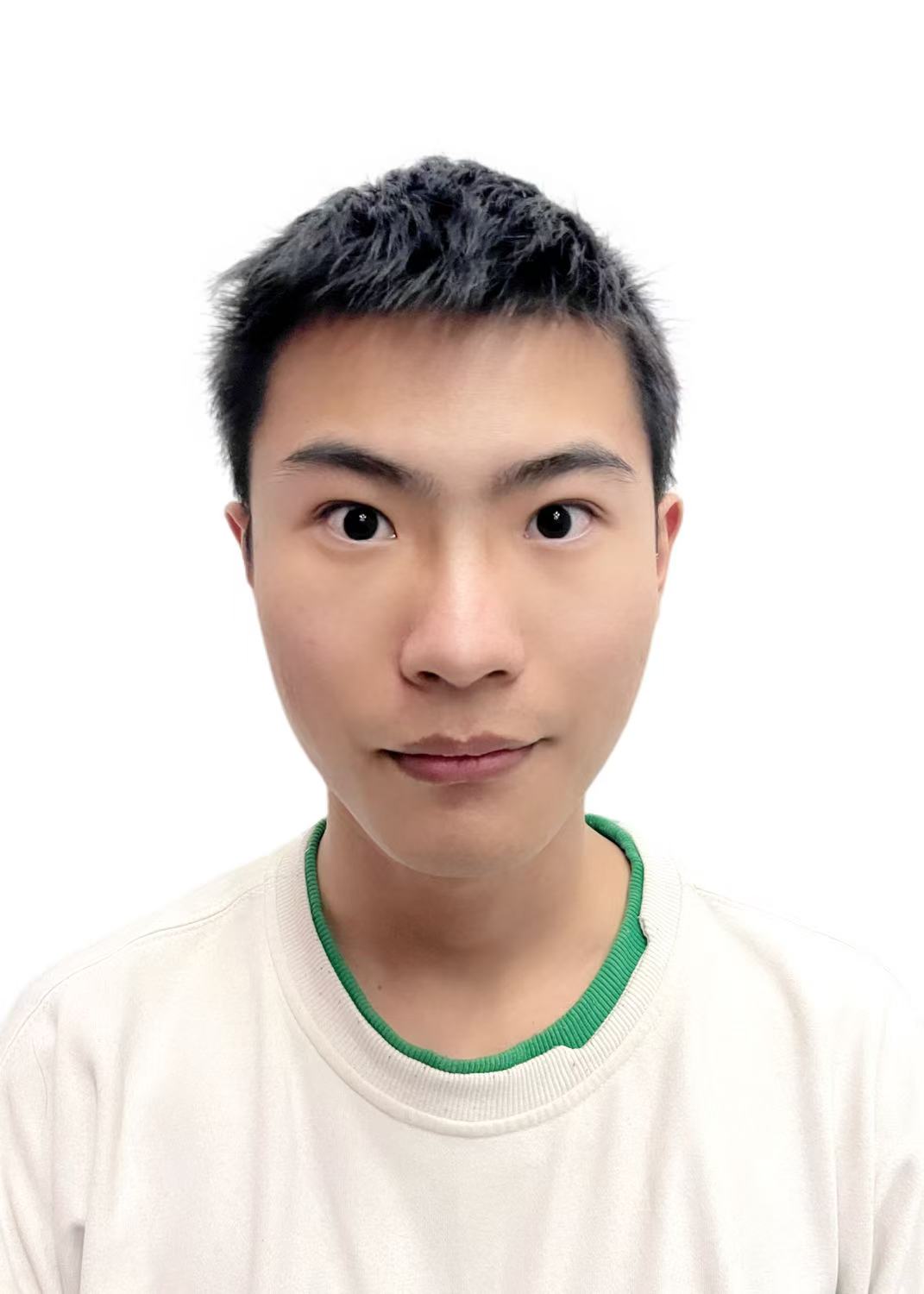}}]{Rui Li} is currently pursuing a Ph.D. degree in Applied Computer Technology at the University of Science and Technology of China. His main research interests include information retrieval and natural language processing. He has published several papers at conferences such as KDD, AAAI, ACL, and EMNLP.
\end{IEEEbiography}
\vspace{-0.55in}
\begin{IEEEbiography}[{\includegraphics[width=1in,height=1.25in,clip,keepaspectratio]{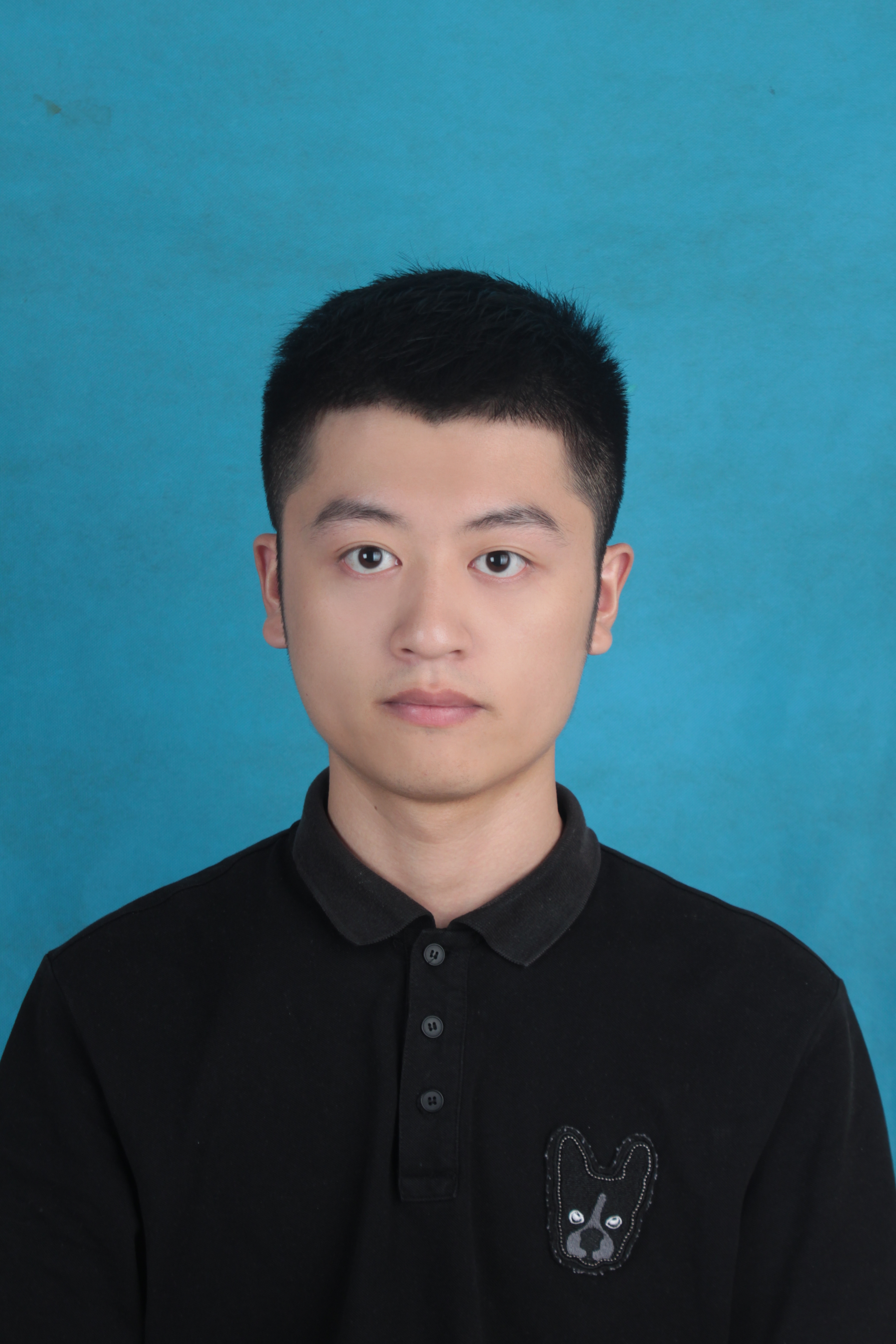}}]{Weibo Gao}   received his BE degree from the School of Software at Hefei University of Technology, China, in 2019. He is currently pursuing a PhD degree in the School of Computer Science and Technology at the University of Science and Technology of China under the supervision of Prof. Qi Liu. He has published several papers in refereed conference proceedings, such as NeurIPS, KDD, SIGIR, TheWebConf, and AAAI. His current research interests include data mining, intelligent education, and generative agents.
\end{IEEEbiography}
\vspace{-0.55in}
\begin{IEEEbiography}[{\includegraphics[width=1in,height=1.25in,clip,keepaspectratio]{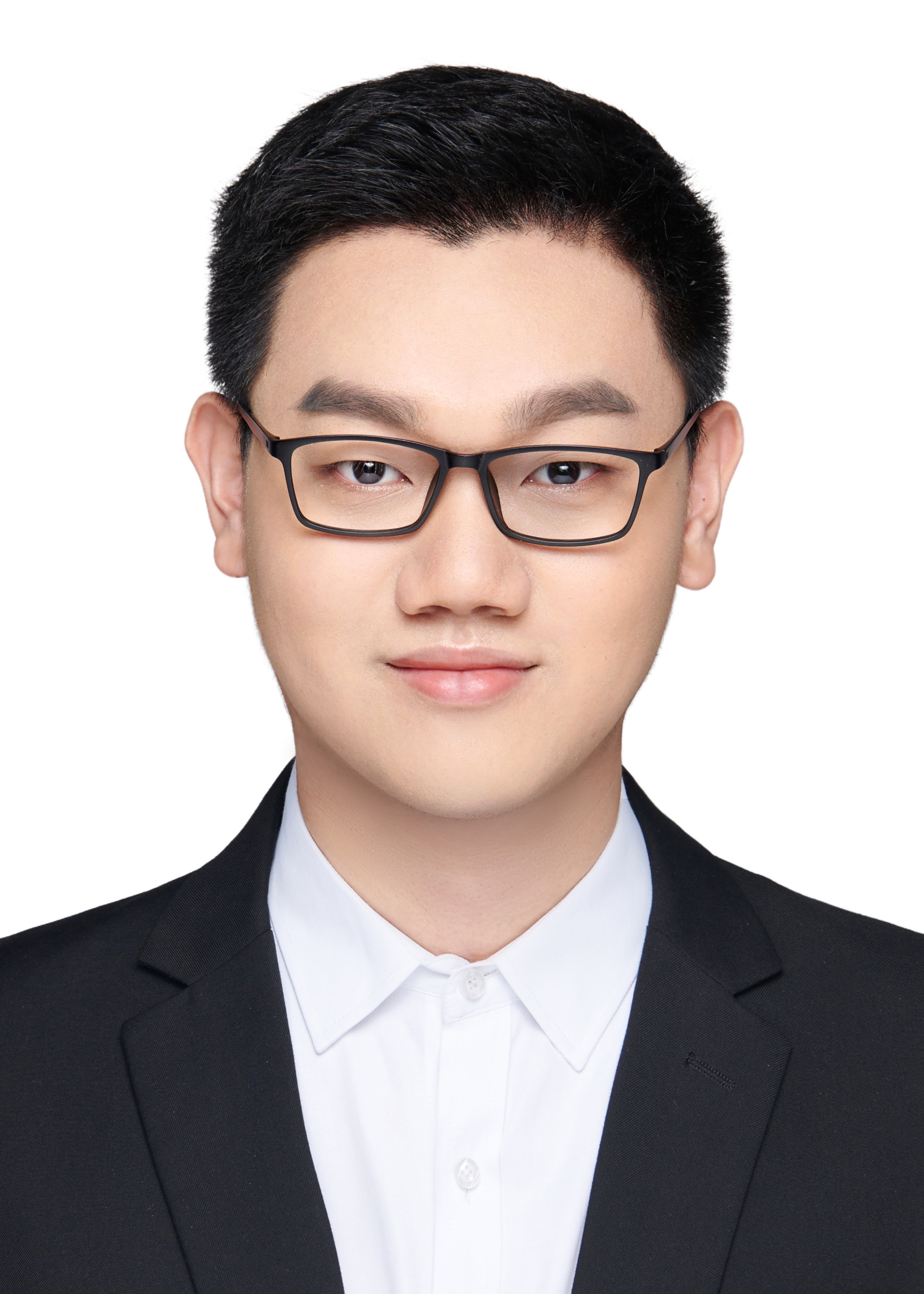}}]{Qingyang Mao}  received the BE degree in statistics from School of Gifted Young at University of Science and Technology of China, Hefei, China. He is currently working toward a Ph.D.  degree in intelligent science and technology with the School of Artificial Intelligence and Data Science. His research interests include cross-domain transfer learning and tabular data mining. He has published several papers in journals and conference proceddings, such as ACM SIGIR, ACM TKDD, WWW, NeurIPS and DASFAA. 
\end{IEEEbiography}
\vspace{-0.55in}
\begin{IEEEbiography}[{\includegraphics[width=1in,height=1.25in,clip,keepaspectratio]{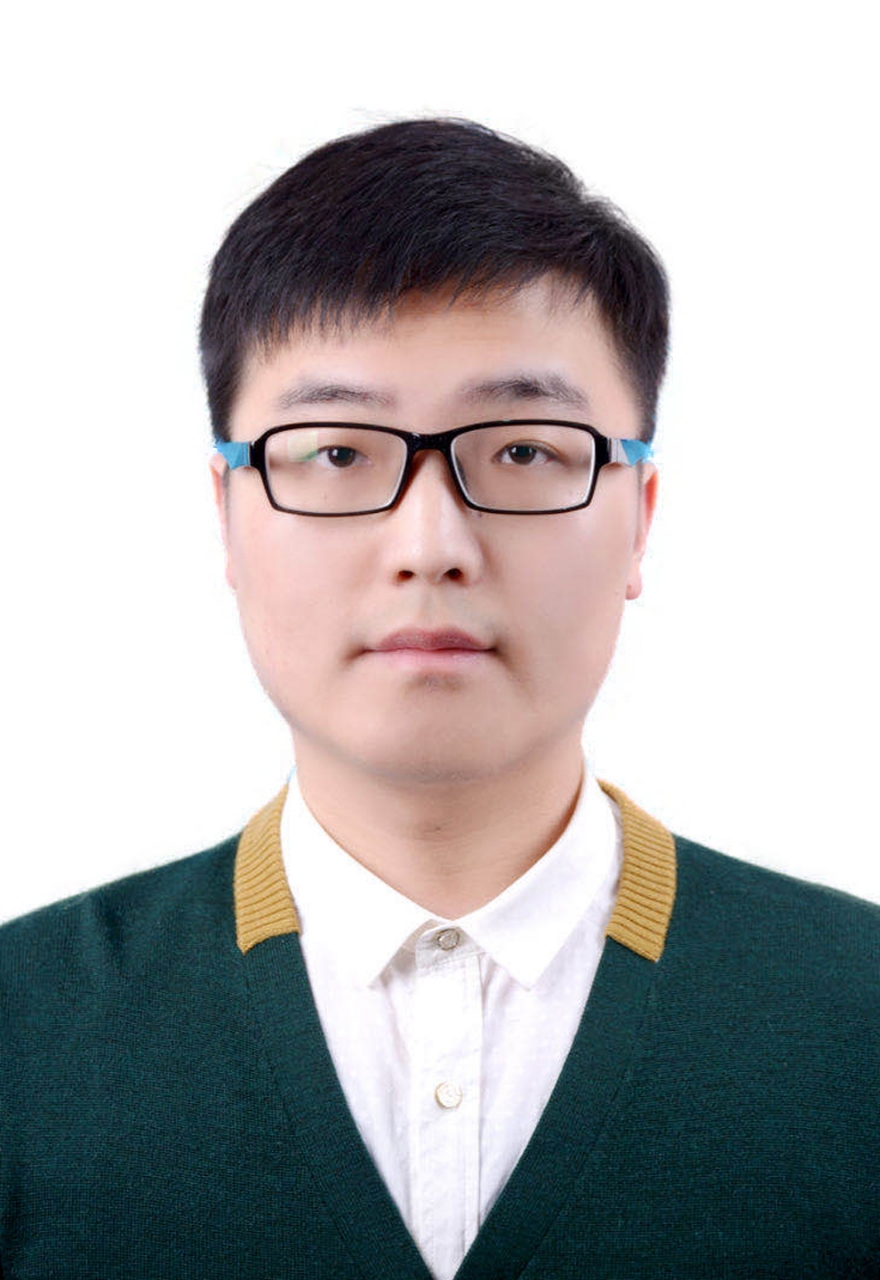}}]{Zhenya Huang}(Member, IEEE)
received the Ph.D. degree from the University of Science and Technology of China (USTC), in 2020. He is currently an Associate Professor with USTC. His main research interests include artificial intelligence, knowledge reasoning, and intelligent education. He has published more than 50 papers in refereed journals and conference proceedings, including TKDE, TOIS, TNNLS, AAAI, KDD, SIGIR, and ICDM. Dr. Huang has served regularly on the program committee of numerous conferences and is a reviewer for the leading academic journals.
\end{IEEEbiography}
\vspace{-0.55in}
\begin{IEEEbiography}[{\includegraphics[width=1in,height=1.25in,clip,keepaspectratio]{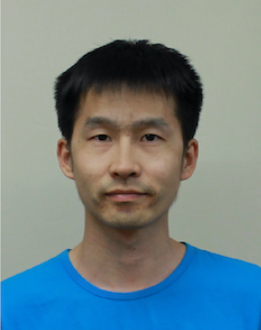}}]{Baosheng Yu} received a B.E. degree from the University of Science and Technology of China (USTC) in 2014 and a Ph.D. degree from the University of Sydney (USYD) in 2019. He is currently a tenure-track assistant professor at Nanyang Technological University, Singapore. His research focuses on machine learning and deep learning, particularly their applications in computer vision and medical image analysis. He has authored or co-authored over 50 publications in leading international conferences and journals, including CVPR, ICCV, ECCV, and IEEE TPAMI.
\end{IEEEbiography}
\vspace{-0.55in}
\begin{IEEEbiography}[{\includegraphics[width=1in,height=1.25in,clip]{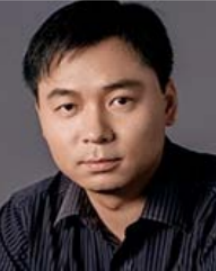}}]{Dacheng Tao (F’15)} is currently a Distinguished University Professor in the College of Computing \& Data Science at Nanyang Technological University. He mainly applies statistics and mathematics to artificial intelligence and data science, and his research is detailed in one monograph and over 200 publications in prestigious journals and proceedings at leading conferences, with best paper awards, best student paper awards, and test-of-time awards. His publications have been cited over 140K times and he has an h-index 180+ in Google Scholar. He received the 2015 and 2020 Australian Eureka Prize, the 2018 IEEE ICDM Research Contributions Award, the 2019 Diploma of The Polish Neural Network Society, and the 2021 IEEE Computer Society McCluskey Technical Achievement Award. He is a Fellow of the Australian Academy of Science, AAAS, ACM and IEEE.
\end{IEEEbiography}
\end{document}